\documentclass[journal]{IEEEtran}

\usepackage[latin9]{inputenc}

\usepackage{times}
\usepackage{epsfig}
\usepackage{graphicx}
\usepackage{amssymb}
\usepackage{cite}
\usepackage{amsmath,amssymb,amsfonts}
\usepackage{algorithmic}
\usepackage{graphicx}
\usepackage{textcomp}
\usepackage{xcolor}
\usepackage{color,soul}
\usepackage{bbm}
\usepackage[font=small,labelfont=bf]{caption}
\usepackage[caption=false,font=small,labelfont=small,textfont=sf]{subfig}
\usepackage{xfrac}

\usepackage[pagebackref=true,breaklinks=true,letterpaper=true,colorlinks,bookmarks=false]{hyperref}

\usepackage{stfloats}

\def \v{\boldsymbol{\mathrm{v}}}
\def \n{\boldsymbol{\mathrm{n}}}
\def \x{\boldsymbol{\mathrm{x}}}
\def \y{\boldsymbol{\mathrm{y}}}
\def \X{\boldsymbol{\mathrm{X}}}
\def \Y{\boldsymbol{\mathrm{Y}}}

\def \p{\boldsymbol{\mathrm{p}}}

\def \I{\boldsymbol{\mathrm{I}}}

\newcommand{\bibliofont}{\small}
\let\OLDthebibliography\thebibliography
\renewcommand\thebibliography[1]{
	\OLDthebibliography{#1}
	\setlength{\parskip}{0pt}
	\setlength{\itemsep}{0pt plus 0.2ex}
	\bibliofont 
}

\renewcommand{\baselinestretch}{1} 

\begin{document}

\title{Modeling the Nonsmoothness of Modern Neural Networks}

\author{
Runze Liu, Chau-Wai Wong, Huaiyu Dai
\thanks{R. Liu, C.-W. Wong and H. Dai are with the Department of Electrical and Computer Engineering, NC State University, Raleigh, NC 27695, USA. (e-mail: rliu10@ncsu.edu; chauwai.wong@ncsu.edu; hdai@ncsu.edu)}
}

\maketitle

\begin{abstract}
Modern neural networks have been successful in many regression-based tasks such as face recognition, facial landmark detection, and image generation.
In this work, we investigate an intuitive but understudied characteristic of modern neural networks, namely, the nonsmoothness.
The experiments using synthetic data confirm that such operations as ReLU and max pooling in modern neural networks lead to nonsmoothness. 
We quantify the nonsmoothness using a feature named the sum of the magnitude of peaks (SMP) and model the input--output relationships for building blocks of modern neural networks. Experimental results confirm that our model can accurately predict the statistical behaviors of the nonsmoothness as it propagates through such building blocks as the convolutional layer, the ReLU activation, and the max pooling layer. We envision that the nonsmoothness feature can potentially be used as a forensic tool for regression-based applications of neural networks.
\end{abstract}

\section{Introduction}

In the past few years, the fast development of convolutional neural networks (CNN) has led to many successful regression-based applications such as face recognition \cite{parkhi2015deep}, facial landmark detection \cite{zhang2014facial}, and 3D pose estimation \cite{mahendran20173d}. CNN has also been used in many other fields such as 2D/3D registration of medical images \cite{miao2016real} and stock price prediction \cite{gudelek2017deep}.

Another successful use of neural networks is face synthesizing.
Deepfake videos synthesized by neural networks nowadays can achieve high quality and look authentic to human eyes\cite{mirsky2020creation}. 
It has raised public concerns because such synthesized images may have detrimental usages. 
For example, deepfake has been used to swap two person's faces in videos, which has posed serious challenges to privacy, civic engagement, and governance. 
Detecting the images and videos synthesized by neural networks has become an emerging research topic and researchers have been exploring different methods to detect these images and videos.
Some approaches exploited biological features \cite{ciftci2020fakecatcher} for deepfake detection. Other methods used artifacts/processing traces left by neural networks for detection \cite{durall2020watch, guarnera2020deepfake, frank2020leveraging}.

In this work, we investigate an intuitive but understudied characteristic, the \textit{nonsmoothness}, of modern neural networks comprising convolutional layers, rectified linear unit (ReLU) activation functions, and max pooling layers.
Our investigation was inspired by the exploration of deepfake generation and detection.
The nonsmoothness can be considered as an artifact of the neural network when it is treated as a processing unit, and we believe that it can be potentially used as a generic forensic tool in many regression-based applications.
We will mathematically define the nonsmoothness and use synthetic data to confirm its existence.
We will study the statistical properties of the nonsmoothness and model the events of nonsmoothness.

The rest of the paper is organized as follows.
In Section~\ref{sec:background}, we review key building blocks of modern neural networks.
We define the concept of nonsmoothness for neural networks in Section~\ref{sec:definition}, and confirm its existence via experiments in Section~\ref{sec:simulation}.
In Section~\ref{sec:modeling}, we model the nonsmoothness events.
In Section~\ref{sec:conclusion}, we conclude the paper.

\section{Background and Preliminaries} \label{sec:background}

\noindent \textbf{Activation Functions}\hspace{3mm}
\begin{figure}[!t]
    \centering
		\vspace{-4mm}
    \subfloat[]{\includegraphics[width=0.465\linewidth]{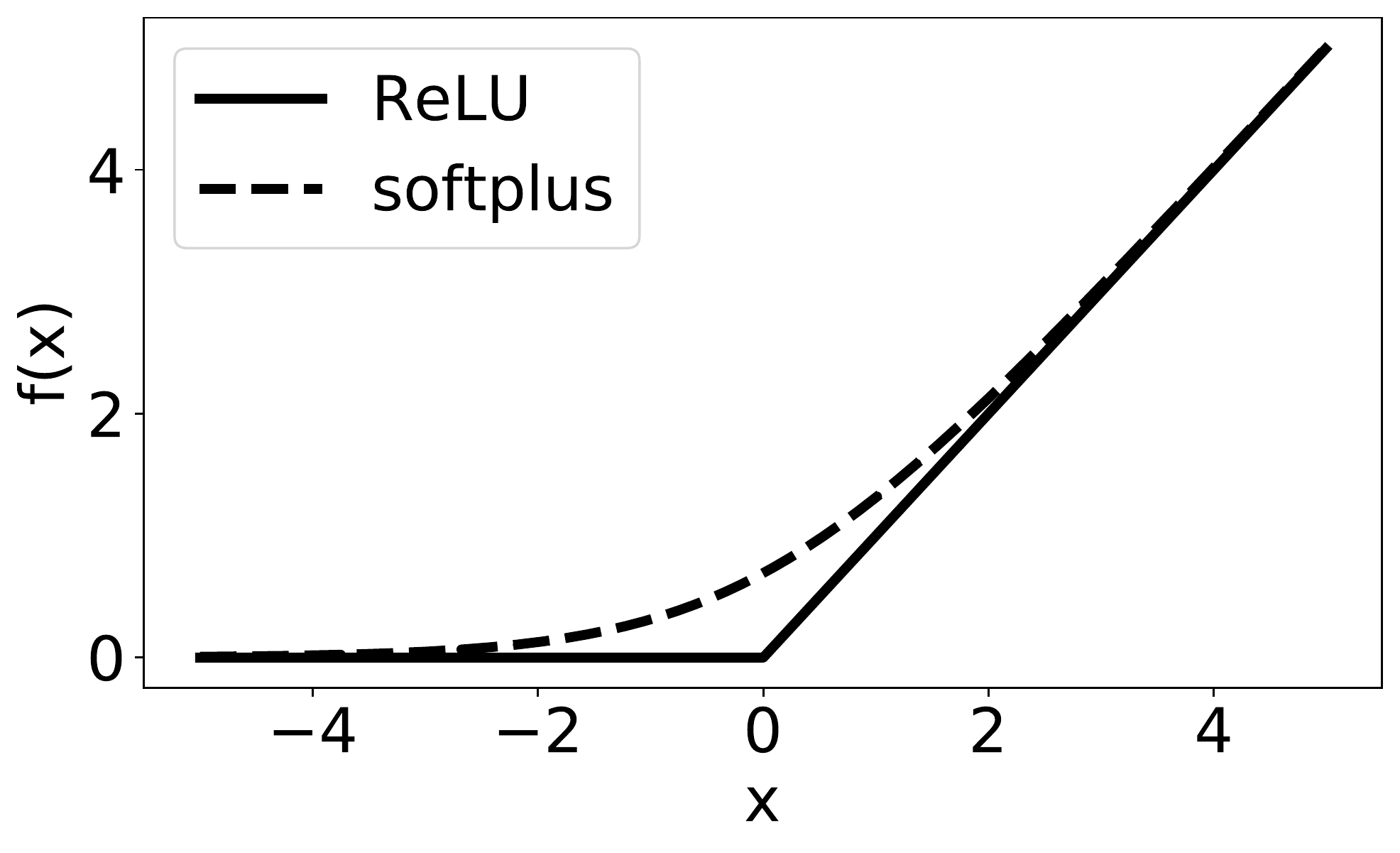}}
    \subfloat[]{\includegraphics[width=0.5\linewidth]{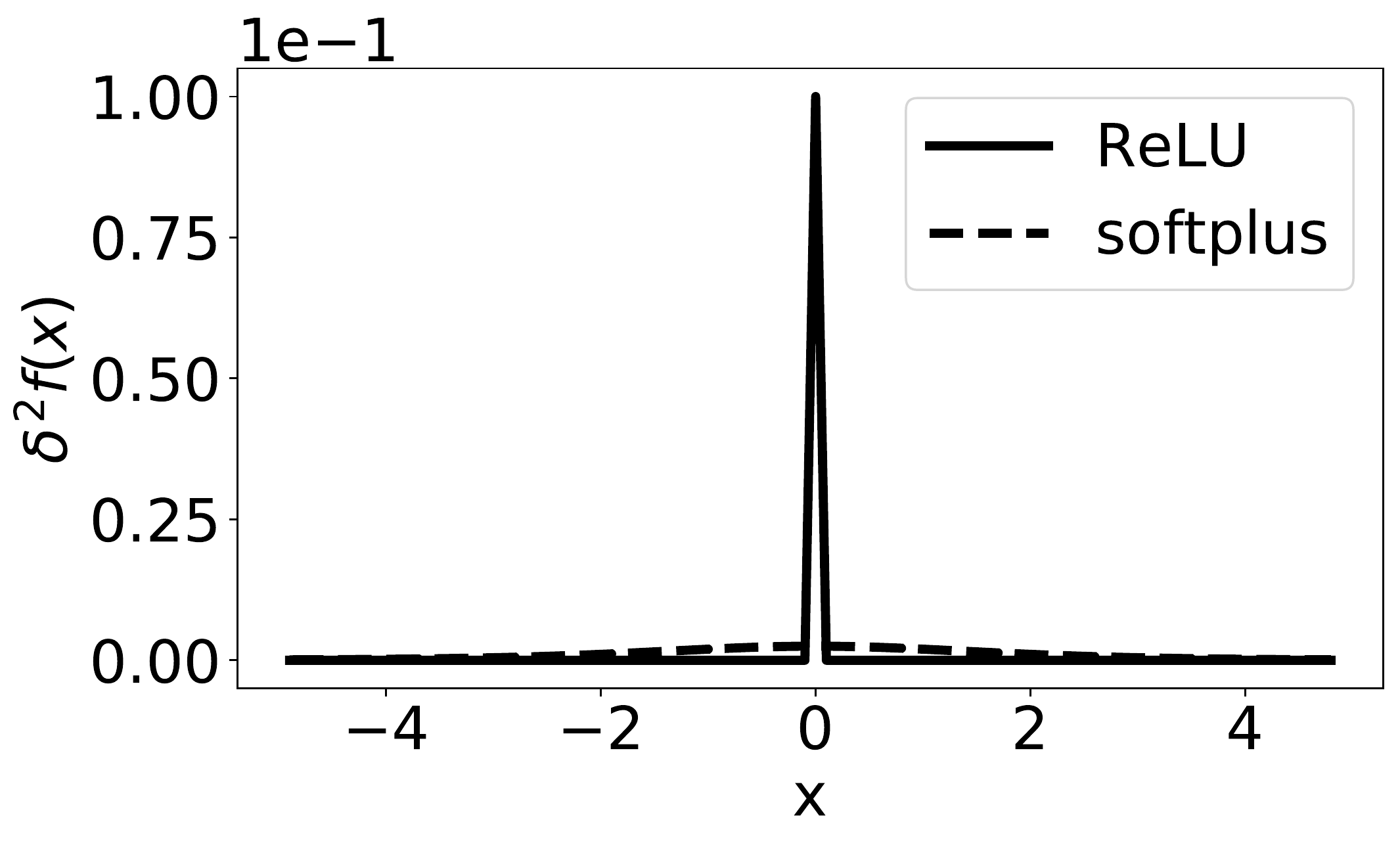}}
    \caption{(a) ReLU and softplus activation functions near $x = 0$. ReLU is nonsmooth at $x = 0$. (b) Second-order difference of ReLU and softplus when the input $x$ is sampled at a step size of $0.1$.}
    \label{relu_softplus}
\end{figure}
Different activation functions have been used in neural networks to introduce nonlinearity. 
Prior to 2011, researchers had mostly used logistic sigmoid or hyperbolic tangent as activation functions.
Glorot et al. \cite{glorot2011deep} argued that the ReLU can better model biological neurons and showed that it can lead to sparse networks.
ReLU is currently widely used for its simple implementation and fast convergence \cite{krizhevsky2012imagenet}. 
ReLU clips negative $x$ to zero while keeping positive $x$ untouched, namely, $g_{\text{ReLU}}(x) = \max(0, x)$. 
The softplus activation \cite{dugas2001incorporating}, $g_{\text{softplus}}(x) = \ln(1 + e^x)$, can be considered as a smooth approximation of ReLU. Fig.~\ref{relu_softplus}(a) compares ReLU and softplus functions. ReLU activation has a change of slope at $x = 0$, while softplus is smooth everywhere.

\noindent \textbf{Pooling Methods}\hspace{3mm}
In convolutional neural networks, pooling layers are used to summarize data via downsampling. The max pooling outputs the largest element within a predefined range, whereas the average pooling outputs the average of all elements. 
Compared with average pooling, max pooling may lead to problems in deep networks because the gradients flow through only the node with maximum value\cite{saeedan2018detail}. Also, Boureau et al. \cite{boureau2010theoretical} showed that max pooling is well suited to sparse features. 
Zhang\cite{zhang2019shiftinvar} discussed the aliasing effect of max pooling and applied anti-aliasing by integrating lowpass filtering to make convolutional networks shift-invariant.
In this work, we will show that ReLU and max pooling introduces nonsmoothness to neural networks, and build mathematical models for the nonsmoothness events.

\section{Nonsmoothness in Modern Neural Networks}\label{sec:definition}

\subsection{Definition of Nonsmoothness}
Given a continuous function $f: \mathbb{R} \rightarrow \mathbb{R}$ and a parameter $t\in \mathbb{R}$, we define that function $f$ is smooth if the first-order derivative, $f'(t) = \underset{h \rightarrow 0}{\lim}\frac{f(t+h) - f(t)}{h}$, exists for all $t$.
For example, $f$ is nonsmooth at $t_0$ when the left-hand and right-hand limits of the first-order derivative do not equal:$\underset{t\nearrow t_0}{\lim}f'(t) \neq \underset{t\searrow t_0}{\lim}f'(t)$.
For simplicity, we denote the left- and right-hand limits at $t_0$ as $f'(t_0^-)$ and $f'(t_0^+)$, respectively.
For the continuous case, a nonsmooth point $t_0$ can be detected if $f'(t_0^+) - f'(t_0^-) \neq 0$. 
If $f(t)$ also contains noise, then by allowing some false positive and false negative, the following detector may be used to detect a nonsmooth point $t_0$:
\begin{equation}
    |f'(t_0^+) - f'(t_0^-)| > \tau_c,
    \label{eq:detect}
\end{equation}
\noindent where $\tau_c$ is a detection threshold.

In the digital world, any input to $f$ is discrete in time, and one intuitive choice for detecting nonsmoothness is to adapt the nonsmoothness detector by replacing the derivative in (\ref{eq:detect}) with the difference operation.
When input $t$ is discrete with a uniform sampling period $\Delta$, i.e., $t \in \Delta \, \mathbb{Z}$, the first-order difference is $\delta f(t) = f(t+\Delta ) - f(t)$, and (\ref{eq:detect}) may be adapted to:
\begin{equation}
    |\delta^2_\Delta f(t)| = |f(t+\Delta) + f(t-\Delta) - 2 f(t)| > \tau_d,
    \label{eq:detect_d}
\end{equation}
\noindent where $\delta^2_\Delta f(t)$ is the second-order difference and $\tau_d$ is a detection threshold for the discrete case.
Similar to (\ref{eq:detect}), there can be false positive and false negative caused by the detector (\ref{eq:detect_d}), but we have to skip the details due to space limitations.
A vector-valued function $f$, i.e., $f(t) = (f_1(t), \dots, \mathcal{N}(t)),$ where $f_i(t) \in \mathbb{R}$ for $i = 1, \dots, n,$ is said to be smooth if $f_i$ is smooth for all $i$'s.

\subsection{Causes of Nonsmoothness in Neural Networks}

\noindent \textbf{Nonsmoothness Caused by Activation Functions}\hspace{3mm}
We show the nonsmoothness caused by activation functions in neural networks.
The neural network will behave nonsmoothly due to the nonsmooth activation functions.
Assuming threshold $\tau_d = 0.02$, ReLU fulfills (\ref{eq:detect_d}) at $0$ but softplus does not, as Fig.~\ref{relu_softplus}(b) shows.
Hence, the detector (\ref{eq:detect_d}) believes that ReLU causes nonsmoothness but softplus does not.

\noindent \textbf{Nonsmoothness Caused by Max Pooling}\hspace{3mm}
\begin{figure}[!t]
    \centering
		\vspace{-5mm}
    \subfloat[]{\includegraphics[width=0.49\linewidth]{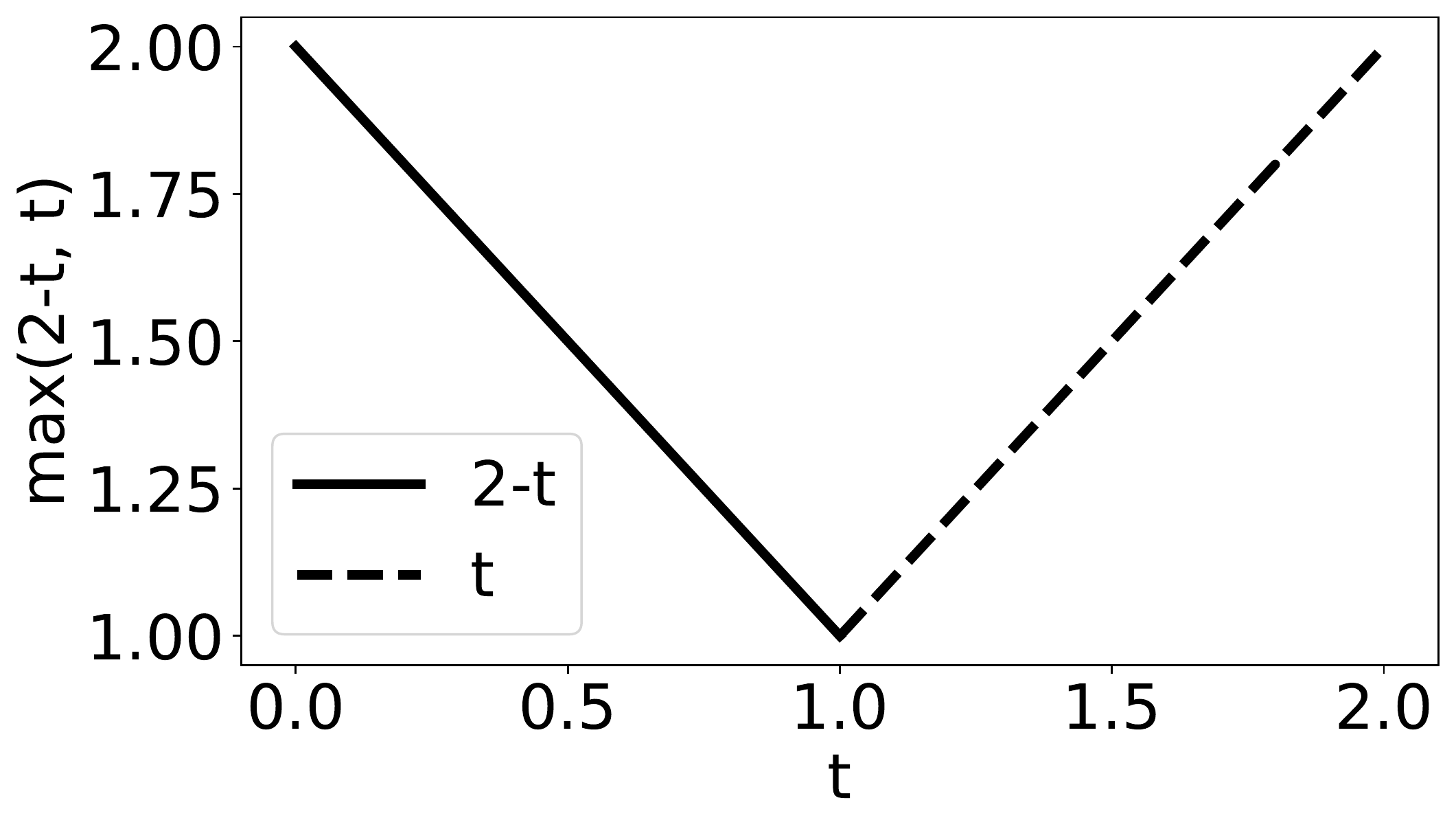}}
    \subfloat[]{\includegraphics[width=0.48\linewidth]{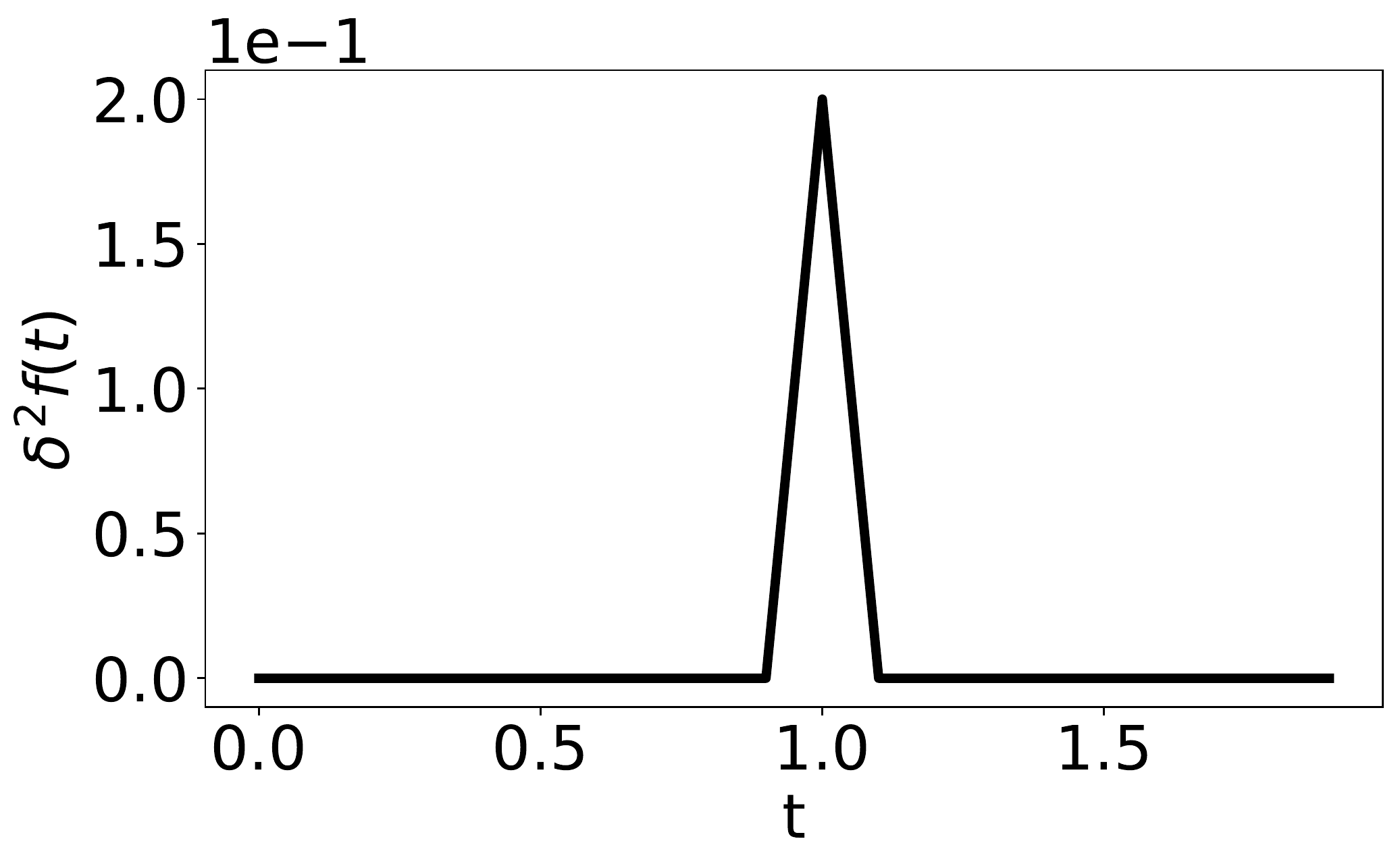}}
    \caption{(a) The output of the max pooling function of a toy example. The curve is nonsmooth at $t = 0$. (b) Second-order difference of the output when $t$ is sampled at a step size of $0.1$.}
		\vspace{-0mm}
    \label{max_pool_toy}
\end{figure}
Max pooling in a convolutional neural network also contributes to nonsmoothness due to the sudden change in the input--output routing as the input changes smoothly. For example, when a max pooling function operates on a smooth $2\times2$ subregion, $\big(x_1(t), x_2(t),x_3(t), x_4(t)\big) = (0,0,2-t,t)$, where the parameter $t$ represents time, the output will be:
\begin{equation}
    \max\big(x_1(t), x_2(t),x_3(t), x_4(t)\big)\!\! =\!\! \left\{
    \begin{array}{ll}
		\!\!\!2 - t ,\!\!\!  &  0 \leq t < 1 ; \\
		\!\!\!t , \!\!\! &  t \geq 1 .
	\end{array}
    \right.
\end{equation}
Note that the output of the max pooling is $2-t$ for $t\in[0,1)$ and $t$ for $t\in[1,\infty)$, as shown in Fig.~\ref{max_pool_toy} demonstrating a sudden routing change at $t=1$ for $x_3(t)$ and $x_4(t)$. The second-order difference is $0.2 > \tau_d$ when the step size is $0.1$, indicating detected nonsmoothness. 
For the same input, the average pooling will instead produce a constant output $\frac{1}{2}$. Compared with the average pooling, the max pooling method causes nonsmoothness in neural networks, and this will be experimentally verified in the next section.

\begin{figure}[!t] 
    \centering
    {\includegraphics[width=0.98\linewidth]{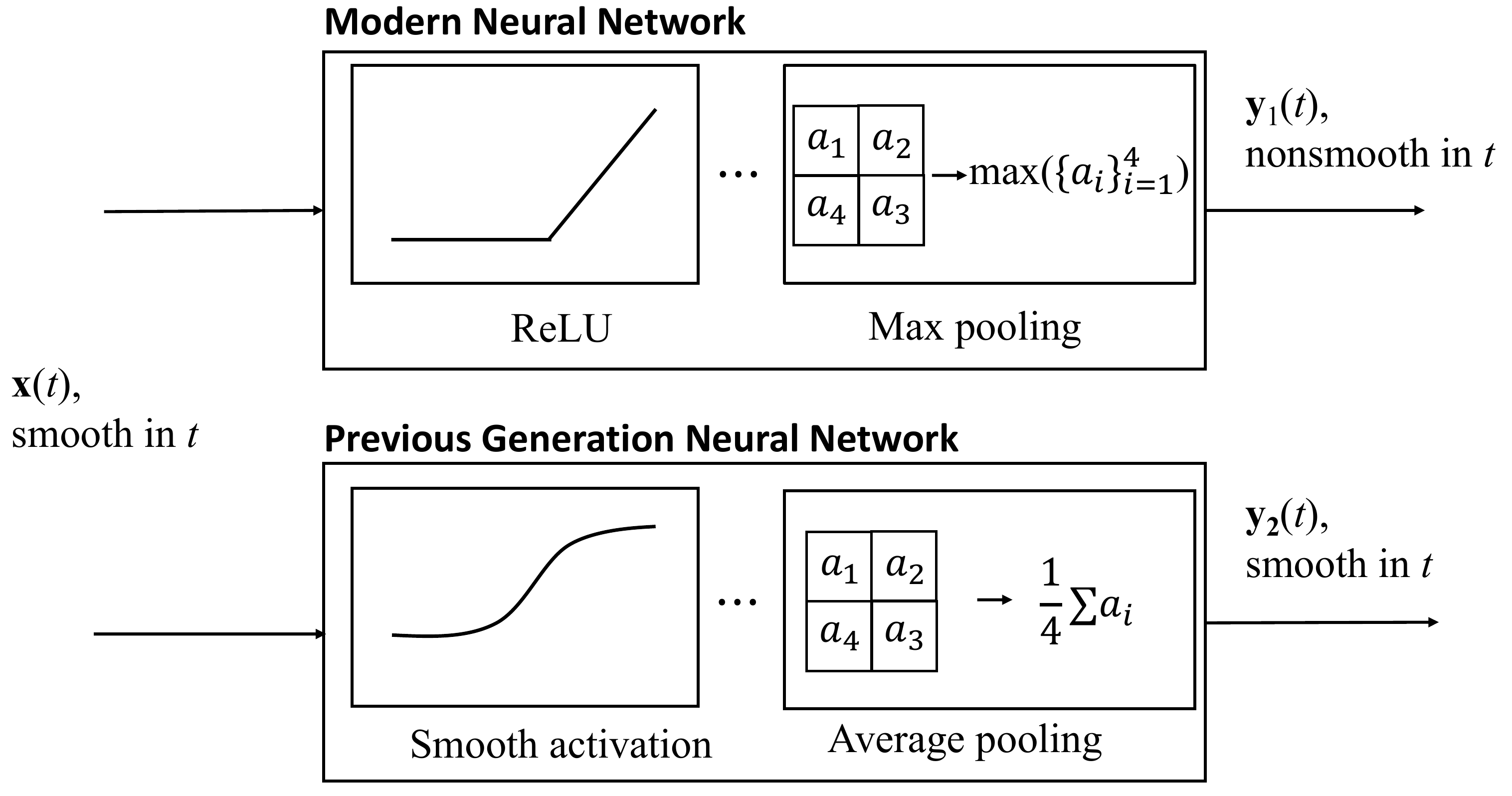}}
    \caption{Modern neural networks prefer nonsmooth ReLU activation and max pooling, whereas previous generation neural networks prefer smooth activations and average pooling. With a smooth input function $\x(t)$ in $t$, output of modern neural network $\y_1(t)$ will be nonsmooth in $t$ due to compositing with nonsmooth functions, whereas output of previous generation neural network $\y_2(t)$ will be smooth.}
    \label{fig:modern_pre_nn}
\end{figure}

Fig.~\ref{fig:modern_pre_nn} summarizes the behaviors of modern and previous generation neural networks.
When the input to the neural network is smooth, modern neural networks with ReLU activation and max pooling will have nonsmooth output, whereas previous generation neural networks \cite{schmidhuber2015deep, cheng1994neural} with smooth activation and average pooling will have smooth output.

\section{Simulated Justification} \label{sec:simulation}

In this section, we use synthetic data to confirm that modern neural networks will cause nonsmoothness.
To illustrate the effect of the nonsmoothness along the time, the input video will be constructed as a smooth function of the time, letting the representation of an image moving smoothly on a manifold.
In the experiments, we will use autoencoders to reconstruct the input videos frame by frame, such that the collocated pixel values in an input video and its reconstructed video can be directly compared for examining the effect of nonsmoothness.
We will also show another example that the representation of an image moving smoothly in the Euclidean space in Section~\ref{sec:mnist} of the Appendix.

\subsection{Dataset Generation and Autoencoder Training}
In the simulation, we let the image vary smoothly such that its representation in the high dimensional space forms a smooth trajectory on a manifold. We used a sphere or part of an ellipsoid to coarsely model a human face under changing illumination. 
We rendered a video of the ellipsoid with a moving point light source. 
The highlight of the ellipsoid changes smoothly as the light moves. 

An ellipsoid is a quadric surface that follows the equation $\frac{x^2}{a^2} + \frac{y^2}{b^2} + \frac{z^2}{c^2} = 1$. 
In this work, we used a half ellipsoid above the $xy$-plane, i.e., $z > 0$ and fix $a = 2.5, b = 4, c = 1$.
To create a smooth video, we illuminate an ellipsoid by moving a point light source at a constant speed. 
We calculate the perceived intensity of a point $\p$ on the ellipsoid by the fully diffuse light reflection model\cite{book}, namely, $\max(\v_i^T  \n, \ 0 )$, where $\v_i$ is the incident light direction at $\p$, and $\n$ is the normal vector at $\p$ on the ellipsoid.

\begin{figure}[!t]
    \centering
		\vspace{-3mm}
    \subfloat[]{\includegraphics[width=0.47\linewidth]{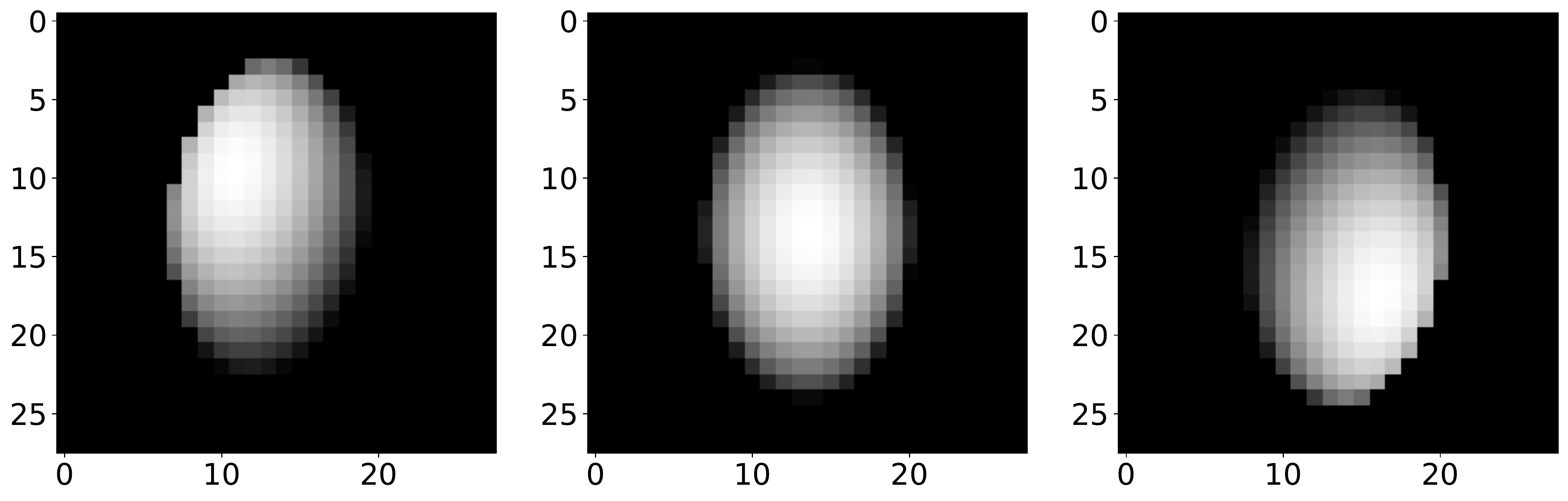}}
    \hspace{4mm}
    \subfloat[]{\includegraphics[width=0.47\linewidth]{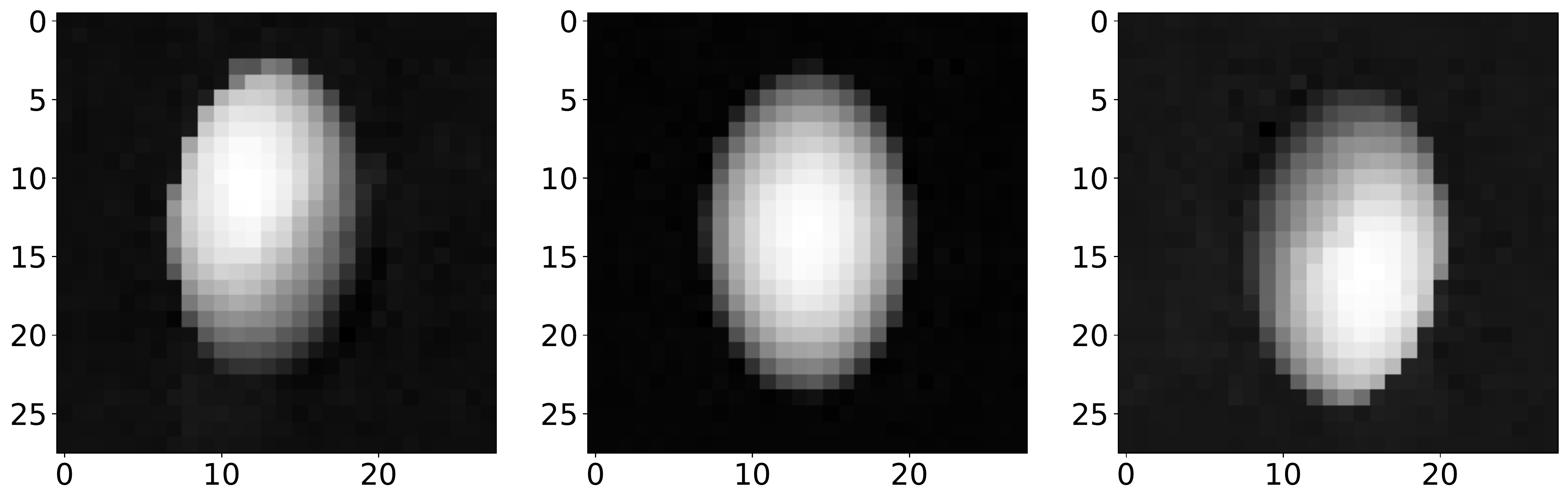}}
    \caption{(a) Three frames of an input video. The point light source moved from northwest to southeast. (b) The reconstructed frames of those in (a) using a trained autoencoder. The reconstructed images are slightly blurred. }
    \label{ellipsoid_frames}
\end{figure}

To construct a training dataset, we randomly generated $10000$ point light source locations $\{(x_i, y_i, 20)\}_{i=1}^{10000}$, where $x_i, y_i \overset{\text{iid}}{\sim}  \mathcal{U}(-10, 10)$. Another $1000$ images were generated in the same way to construct the validation dataset. 
Several ellipsoid images with different light source locations are shown in Fig.~\ref{ellipsoid_frames}(a).

Two autoencoders were trained to reconstruct the ellipsoid images of different light source locations. 
One autoencoder follows Setup~1: ReLU activation and max pooling, and the other autoencoder follows Setup~2: softplus activation and average pooling. 
Adam optimizer and MSE loss were used for training. The learning rate was $10^{-3}$ to ensure a fast training speed and good training performance for the scale of this problem. 
To avoid overfitting, the model with the lowest validation error is considered as the trained model.
The reconstructed frames using the trained autoencoder are shown in Fig.~\ref{ellipsoid_frames}(b).

\subsection{Processing Smooth Input Videos by Neural Networks}
\begin{figure}[!t]
    \centering
    \subfloat[]{\includegraphics[width=0.49\linewidth]{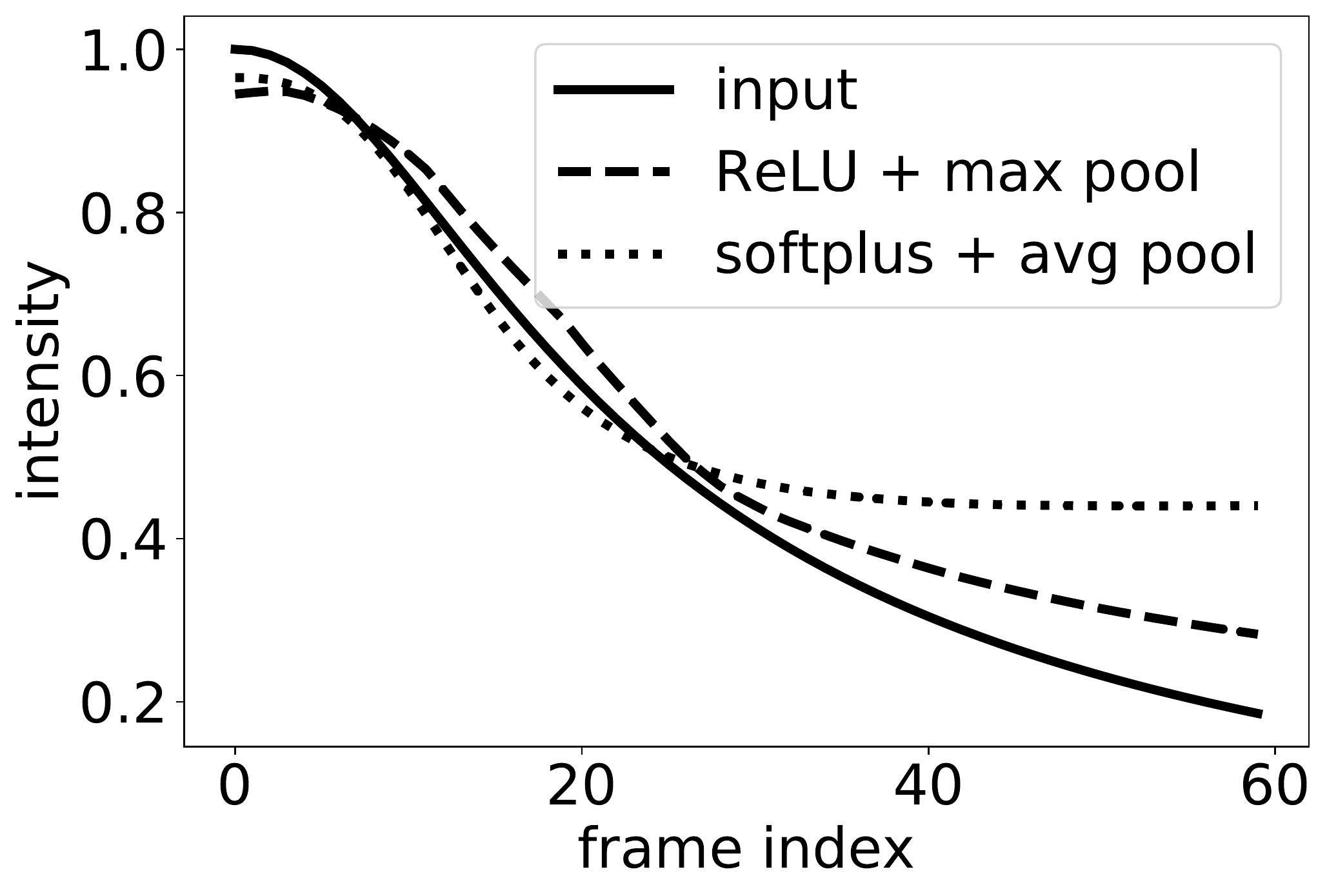}}
    \subfloat[]{\includegraphics[width=0.49\linewidth]{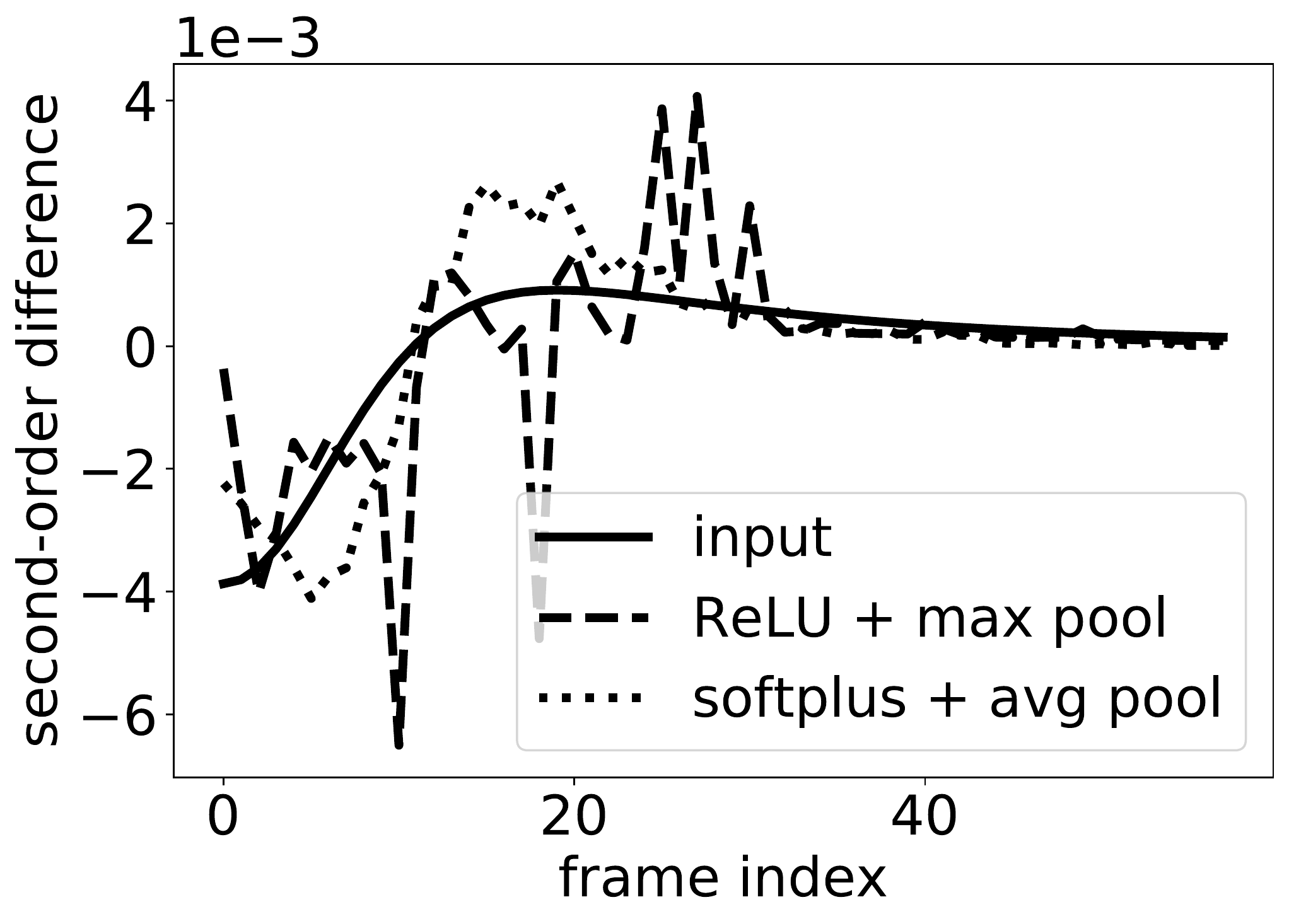}}
    \caption{For a representative pixel location, (a) intensity curves and (b) second-order difference curves of input video and reconstructed videos using the autoencoders of \textsc{ReLU+MaxPool} and \textsc{Softplus+AvePool}. The autoencoder of \textsc{ReLU+MaxPool} caused more nonsmoothness in terms of the second-order difference.  }
    \label{ellipsoid_curves}
\end{figure}

We show the nonsmoothness introduced by the autoencoder. 
First, we need an input video that is smooth in time. We constructed a test video where the light source moves from $(-9, -9, 20)$ to $(9, 9, 20)$ at a speed of $2$ pixels per second. 
We reconstructed the test video with the two trained autoencoders, \textsc{ReLU+MaxPool} or \textsc{Softplus+AvePool}. 
For a pixel location, the intensity curves of the input video and reconstructed videos with two autoencoders are shown in Fig.~\ref{ellipsoid_curves}(a). 
The trends of the curves appear smooth because the goal of an autoencoder is to reconstruct the input video.
The second-order difference curves are shown in Fig.~\ref{ellipsoid_curves}(b). Compared to the autoencoder using \textsc{Softplus+AvePool}, the autoencoder using \textsc{ReLU+MaxPool} has introduced more nonsmoothness in terms of second-order difference.

\begin{figure}[!t]
    \centering
		\vspace{-4mm}
    \subfloat[]{\includegraphics[width=0.49\linewidth]{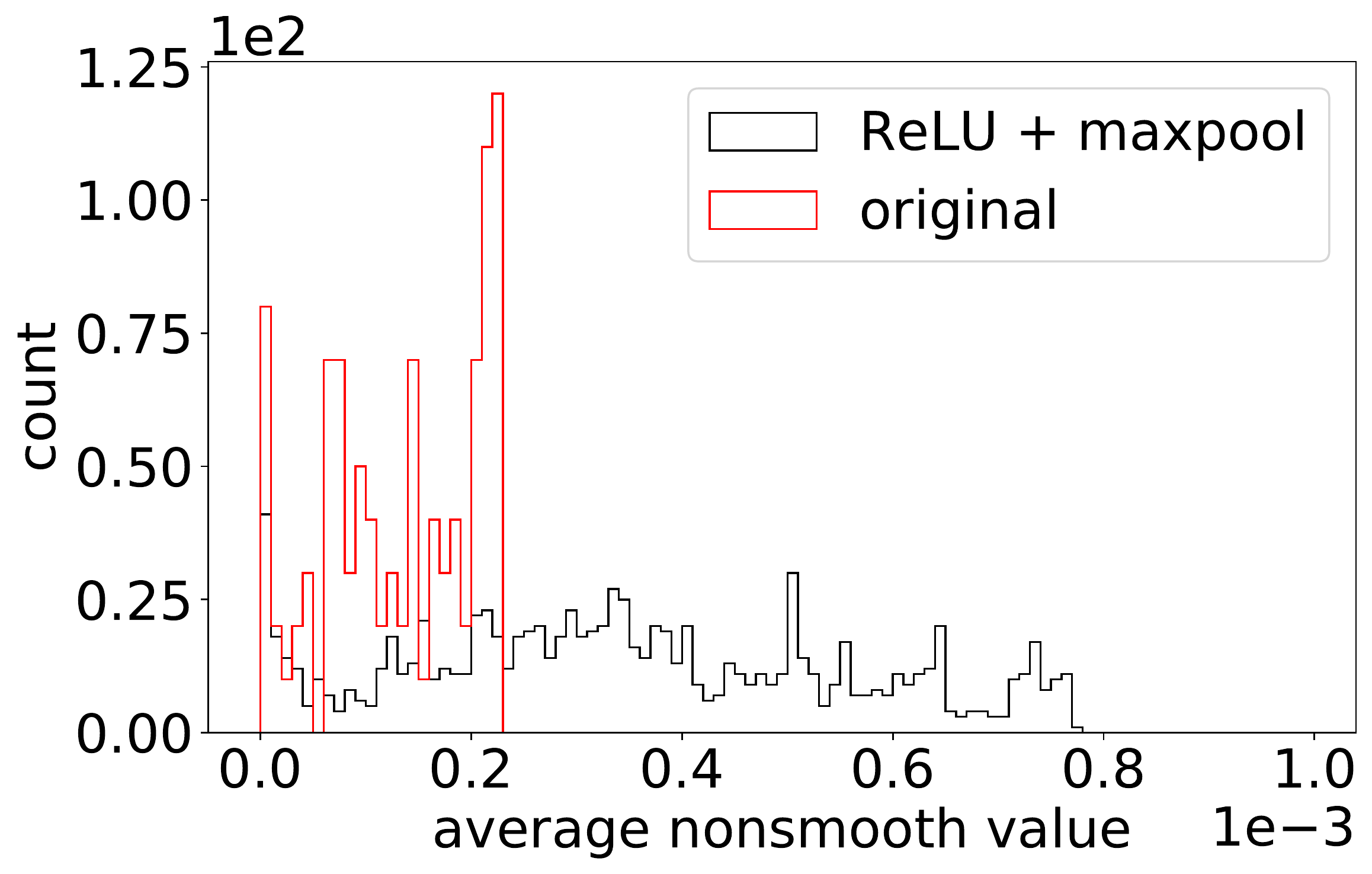}}
    \subfloat[]{\includegraphics[width=0.49\linewidth]{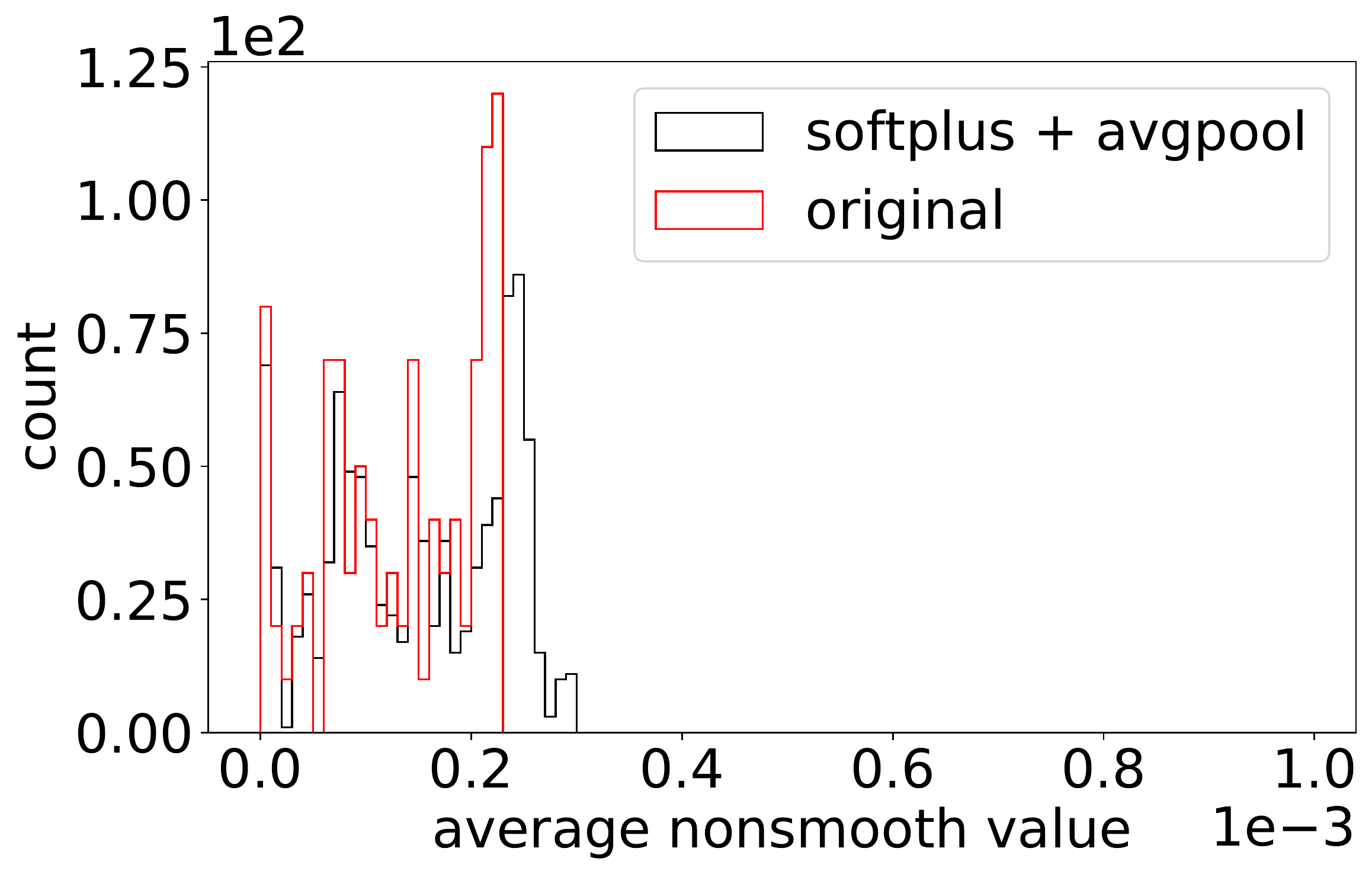}}
    \caption{The histograms of the AveNonSmooth of the  original and the reconstructed videos from (a) \textsc{ReLU+MaxPool}, and (b) \textsc{Softplus+AvePool}. When using autoencoder with \textsc{ReLU+MaxPool} to reconstruct videos, the AveNonSmooth is larger in general.}
    \label{ellipsoid_stat_res}
\end{figure}

To quantify the nonsmoothness caused by autoencoders, we define a statistic for a given video named \textit{the average nonsmoothness} as follows: 
\begin{equation}
    \text{AveNonSmooth} = \frac{1}{MNT}\sum_{i=1}^{M}\sum_{j=1}^{N}\sum_{t=1}^{T} |\delta^2y_{i,j}(t)|,
    \label{eq:avg_nonsmooth}
\vspace{-0mm}
\end{equation}
\noindent where $\delta^2y_{i,j}(t)$ is the second-order difference of pixel location $(i, j)$ at time $t$, $M$ and $N$ are the number of rows and columns of the frame, respectively, and $T$ is the total number of frames of the video. 
We examined the distribution of AveNonSmooth of $100$ input videos with random moving paths of the light source reconstructed with ten trained autoencoders.
Fig.~\ref{ellipsoid_stat_res} shows the histograms of the AveNonSmooth of the reconstructed videos. 
\textsc{Softplus+AvePool} hardly increased AveNonSmooth but \textsc{ReLU+MaxPool} increased significantly. Hence, we confirm that the autoencoder with \textsc{ReLU+MaxPool} leads to nonsmoothness.

\section{Modeling of Nonsmoothness Events  } \label{sec:modeling}
\subsection{Motivation for Modeling}
\begin{figure}[!t]%
    \centering
    {\includegraphics[width=0.98\linewidth]{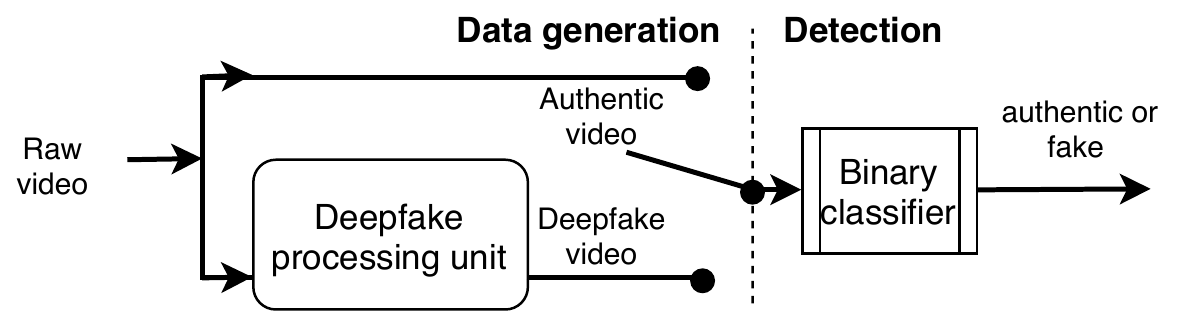}} 
    \vspace{-0mm}
    \caption{Conceptual diagram for authentic/fake videos generation and deepfake detection. The binary classifier aims at detecting the processing unit used to generate deepfake videos.}
		\vspace{-0mm}
    \label{fig:diagram_detection}
\end{figure}

In Sections \ref{sec:definition} and \ref{sec:simulation}, we show that ReLU and max pooling are the causes of nonsmoothness in  neural networks. Such nonsmoothness traces due to neural networks may be used as a forensic tool for regression-based applications. 
For example, when neural networks are used to generate deepfake videos, they can introduce nonsmoothness in the resulting videos, as if an extra processing unit was applied to an authentic video, as shown in the left half of Fig.~\ref{fig:diagram_detection}. A classifier can be exploited to detect this extra processing unit by examining the statistical traces it leaves behind, effectively detecting deepfake videos.
In this section, we investigate and model the statistical behaviors of nonsmoothness events by examining the propagation of nonsmoothness events through various building blocks of neural networks. In this paper, we focus on the analysis of the convolutional layer, ReLU activation, max pooling layer, and the combined effect of a convolutional layer with ReLU activation and a max pooling layer. The modeling results for the transpose convolutional layer are reported in Section~\ref{sec:model_transconv} of the Appendix. We leave the combined effect of all layers of the neural network for future work.

\subsection{Modeling of Convolutional Layers} \label{sec:model_conv}
When an input video is smooth, a large value in the second-order difference curve represents nonsmoothness caused by the neural network. 
In this section, we define a \emph{peak} to be a value that is ten times larger than the sample mean and median in a time series. The occurrence of a peak in the second-order difference curve is said to be a \textit{nonsmoothness event}.
Besides the peaks, the second-order difference for a smooth function may not be zero due to the sampling period that is not infinitesimal. Using the sum of the second-order difference in a time series as in (\ref{eq:avg_nonsmooth}) may overestimate the nonsmoothness and therefore result in inaccuracy in modeling.
To better quantify the nonsmoothness for modeling, we propose to use the \textit{sum of the magnitude of peaks (SMP)} of the second-order difference curve.

\noindent \textbf{Properties of Nonsmoothness Events} \hspace{3mm} To model the nonsmoothness events in neural networks, we first need to investigate their behaviors.
It is important to understand the properties of the nonsmoothness events in the summation of the time series because a convolutional layer in the neural network will linearly combine coordinates of the input data, i.e., the output of convolutional layer is the weighted sum of input nodes.
Given two continuous functions $f_1, f_2: \mathbb{R} \rightarrow \mathbb{R}$, $f_1$ is nonsmooth at $t_1$, i.e., $f_1'(t_1^+) \neq f_1'(t_1^-)$, and $f_2$  is nonsmooth at $t_2$, and define $ f_{\text{sum}}(t) = f_1(t) + f_2(t)$. 
Below, we present two properties of nonsmoothness events.
First, the occurrence times of the nonsmoothness events are inherited, i.e., $ f_{\text{sum}}(t) $ is nonsmooth at $t_1$ and $t_2$. Second, the change of first-order derivatives is inherited, i.e., $ f_{\text{sum}}'(t_1^+) - f_{\text{sum}}'(t_1^-) = f_1'(t_1^+) - f_1'(t_1^-), f_{\text{sum}}'(t_2^+) - f_{\text{sum}}'(t_2^-) = f_2'(t_2^+) - f_2'(t_2^-)$.  
For the discrete case, nonsmoothness events may happen simultaneously in different time series. 
Summing up the magnitude of these peaks will overestimate the nonsmoothness, and a discounting multiplicative factor can be used to compensate for this issue.
In this section, we assume that the occurrence times of nonsmoothness events are different and the second-order differences will be inherited since we have observed from data that the events are rare.

\noindent \textbf{Linear Model for Convolutional Layers} \hspace{3mm} 
We now model the nonsmoothness events in the convolutional neural networks. 
First, we investigate SMP for input nodes and output nodes of a convolutional layer.
Denote the input of the convolutional layer to be $\X \in \mathbb{R}^{ M \times M \times C_{\text{in}} }$, where $C_{\text{in}}$ is the number of input channels and $M \times M$ is the size of the image in each input channel.
Denote the output of the convolutional layer to be $\Y \in \mathbb{R}^{ N \times N \times C_{\text{out} }}$, where $C_{\text{out}}$ is the number of input channels and $N \times N$ is the size of the image in each output channel.
For the convolutional layer, denote the size of the convolutional kernel to be $k \times k$, the stride to be $s$, and the padding to be $p$.

For pixel location $(u, v)$ in channel $c$ of input data, denote its SMP by $X_{cuv}$. Similarly, for pixel location $(m, n)$ in channel $c$ of output data, denote its SMP by $Y_{cmn}$. Due to the inheritance of the nonsmoothness events, a linear model can be established for $Y_{cmn}$ and $X_{cuv}$ as follows:
\begin{equation}
	Y_{cmn} = \sum_{c=1}^{C_{\text{in}}} \sum_{u=ms-p}^{ms-p+k-1} \sum_{v=n s-p}^{ns-p+k-1} |W_{cij}| \cdot X_{cuv},
	\label{conv_model}
\end{equation}
\noindent where $W_{cij}$ is the weight in the convolutional kernel for location $(u, v)$ in channel $c$ of input, i.e., $i = u - (ms - p), j = v - (ns - p)$.
Since the channels in the input or output are the same without loss of generality, SMPs in different channels $\{ Y_{cmn} \}_{c=1}^{C_{\text{out}}}$ are identically distributed. Therefore, we define expectations of SMPs to be $\mu^Y_{mn} = \mathbb{E}\left[Y_{cm n}\right]$ for all $c$, and similarly $\mu^X_{mn} = \mathbb{E}\left[X_{cuv}\right]$.
Taking the expectations of both sides of (\ref{conv_model}), the SMP of input and output can be related as follows:
\begin{equation}
	\mu^Y_{mn} =w_0 \cdot C_{\text{in}} \cdot \sum_{u=ms-p}^{ms-p+k-1} \sum_{v=n s-p}^{ns-p+k-1} \mu^X_{mn},
	\label{conv_model_expectation}
\end{equation}
\noindent where we have invoked the assumptions that $\mathbb{E}\left[W_{cij}\right]=w_0 $ for all $(i, j)$ and $c$, and $W_{cij}$ is independent of $X_{cuv}$. The latter assumption is justified by our analysis that the correlation between the weights and the magnitudes of peaks of the real data in Section \ref{sec:simulation} is small, i.e., $0.16$.

We used the simulated ellipsoids data from Section \ref{sec:simulation} to test the effectiveness of the model for nonsmoothness events.
We recorded intensity curves along the time for each node in the autoencoders of \textsc{ReLU+MaxPool} and then calculated SMP for the input nodes and output nodes for the second convolutional layer. 
We did not use the first convolutional layer because the input video is smooth and the SMP of input will be zero.
The sample mean of the SMP at different channels was used to approximate the expectation of SMP, i.e.,  $ \hat{\mu}^Y_{mn} = 1/C_{\text{out}} \sum_{c=1} ^{C_{\text{out}}} y_{cmn}, \hat{\mu}^X_{mn} = 1/C_{\text{in}} \sum_{c=1} ^{C_{\text{in}}} x_{cuv}$, where $ y_{cmn} $  is the SMP at location $(m, n)$ in channel $c$ at the output data of the convolutional layer, and $ x_{cuv} $ is the SMP at location $(u, v)$ in channel $c$ in the input data. $ C_{\text{in}} $ and $ C_{\text{out}} $ are the number of channels in the input and output data, respectively.

\begin{figure*}[!t]
	\vspace{-3mm}
	\subfloat[]{\includegraphics[width=0.2\linewidth]{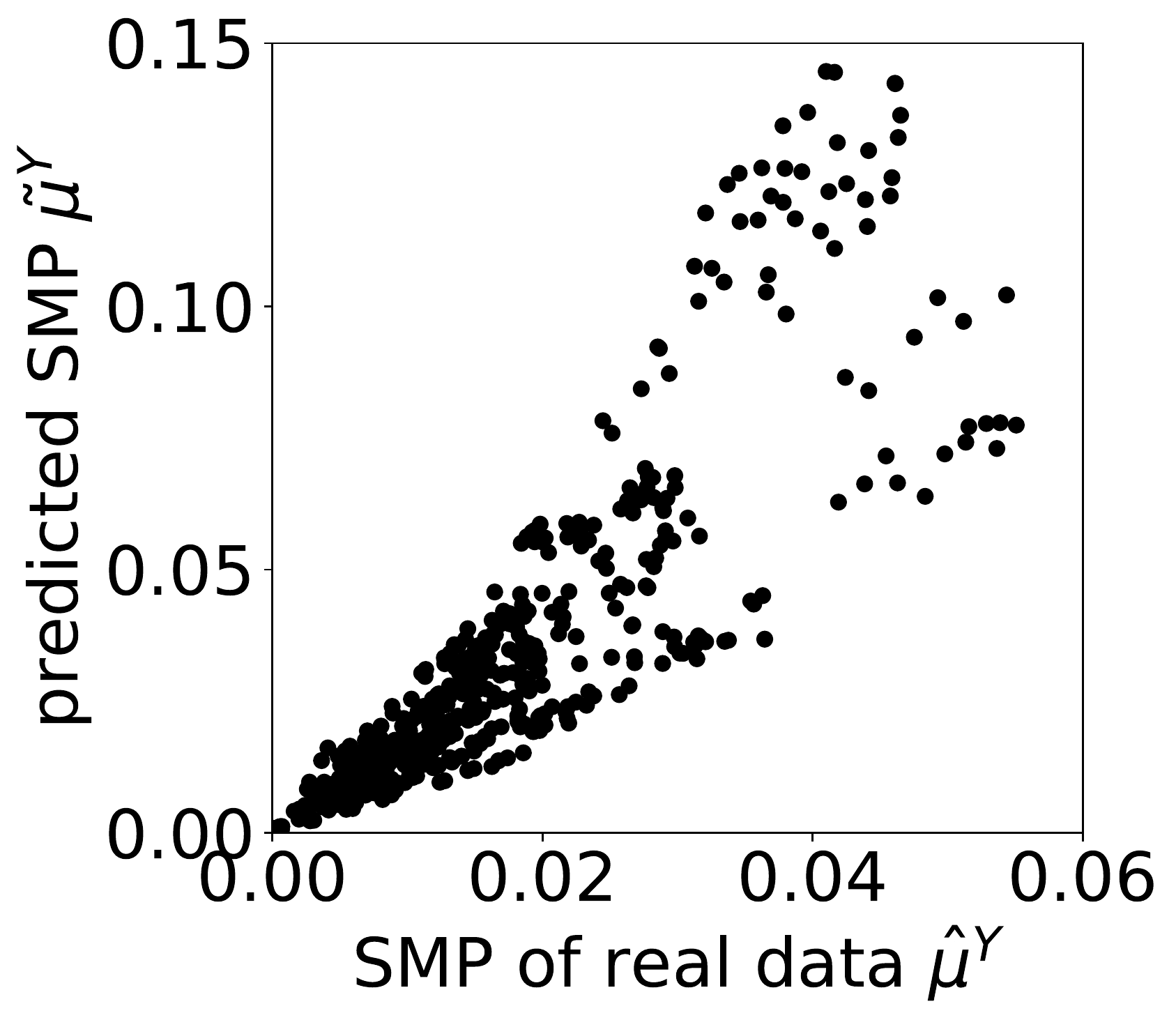}}
	\subfloat[]{\includegraphics[width=0.2\linewidth]{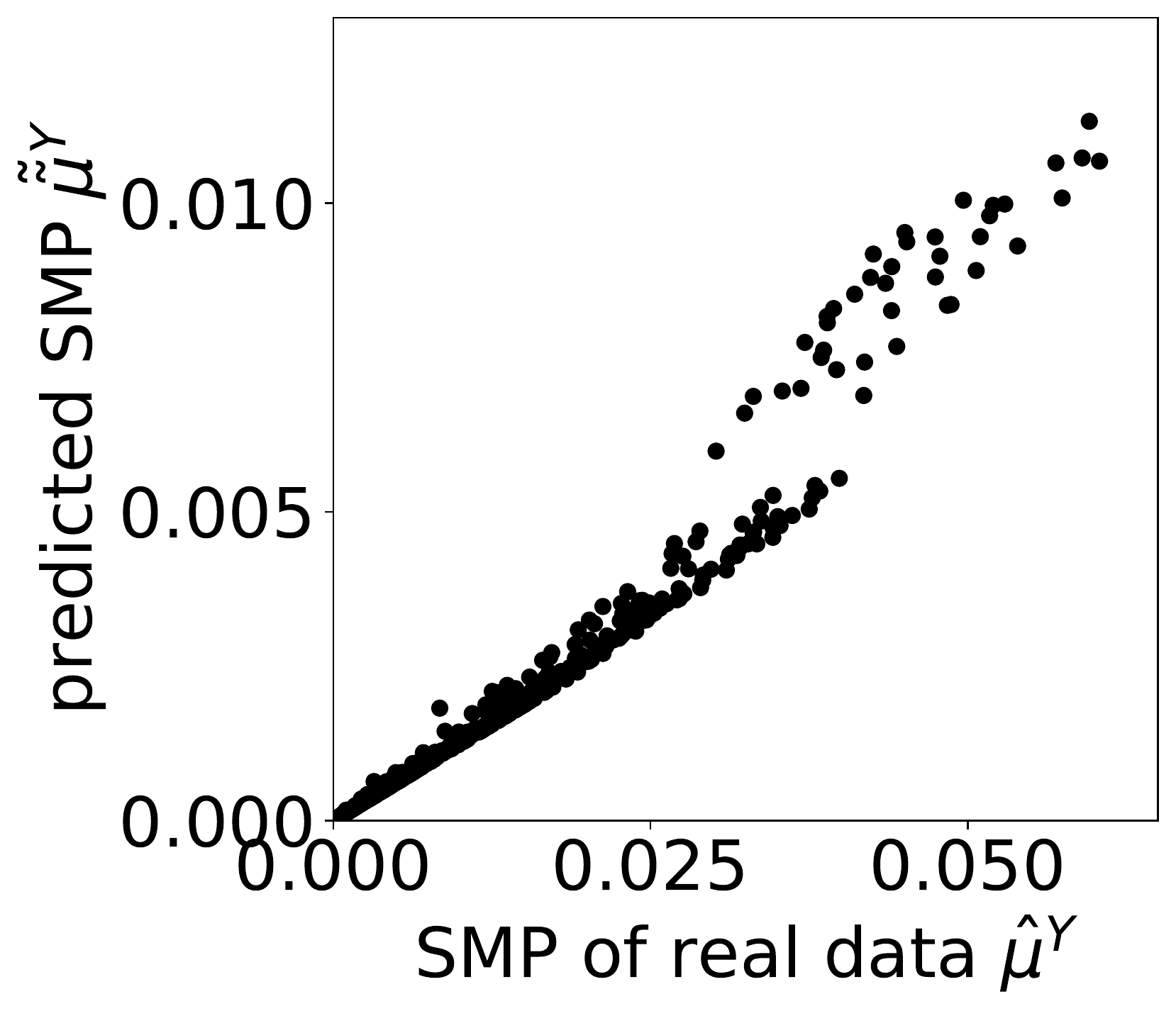}}
	\subfloat[]{\includegraphics[width=0.21\linewidth]{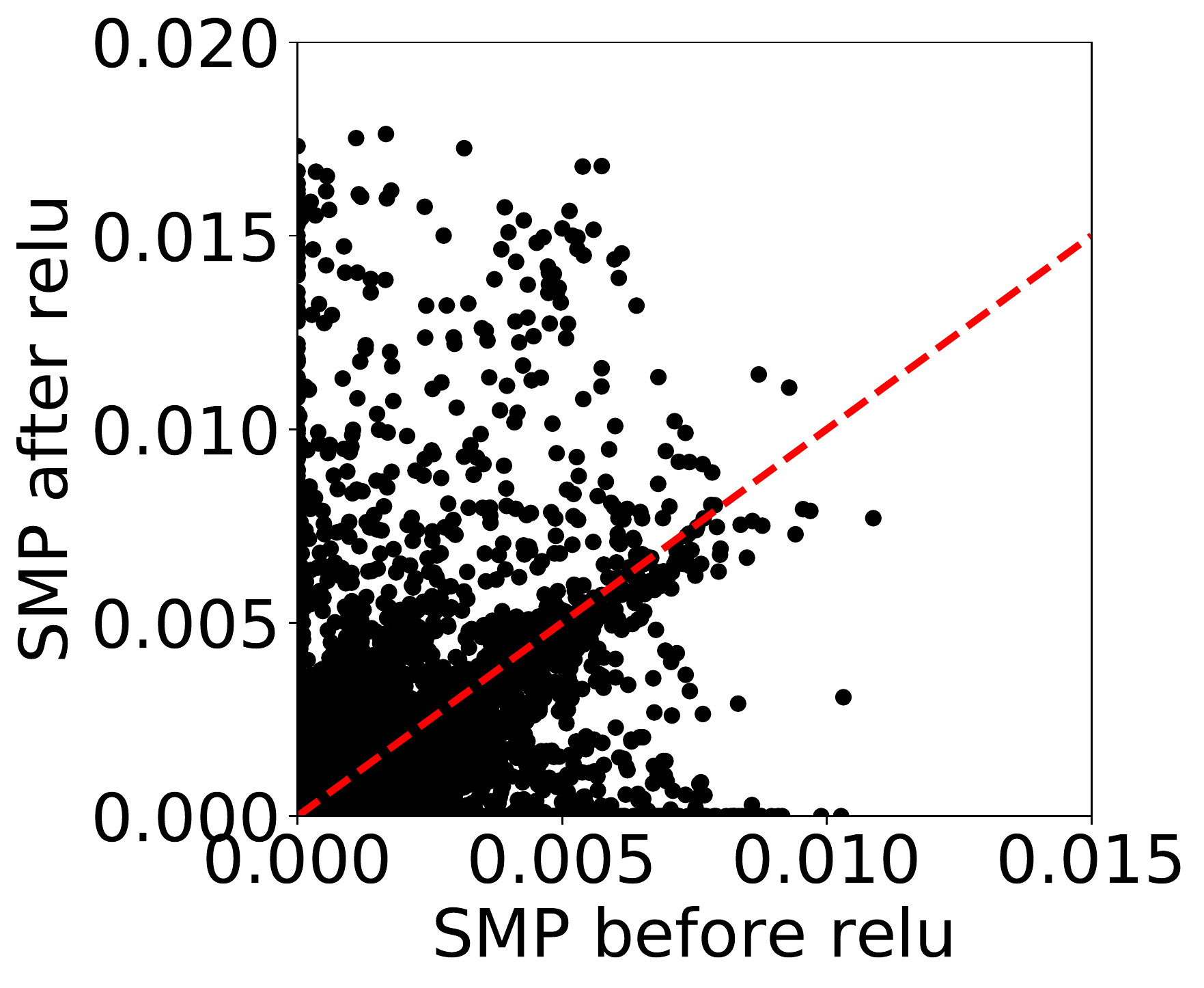}}
	\subfloat[]{\includegraphics[width=0.21\linewidth]{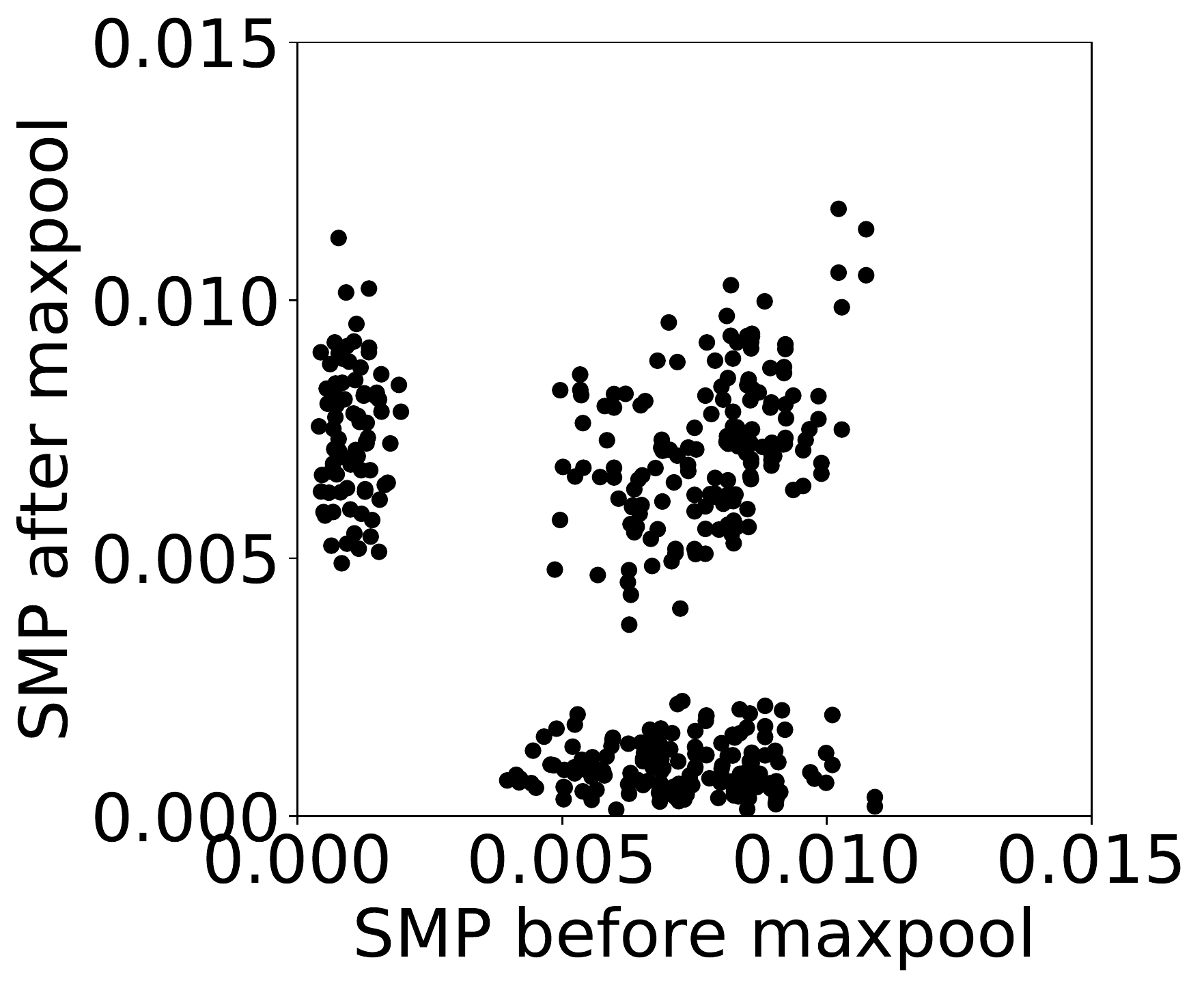}}
	\subfloat[]{\includegraphics[width=0.19\linewidth]{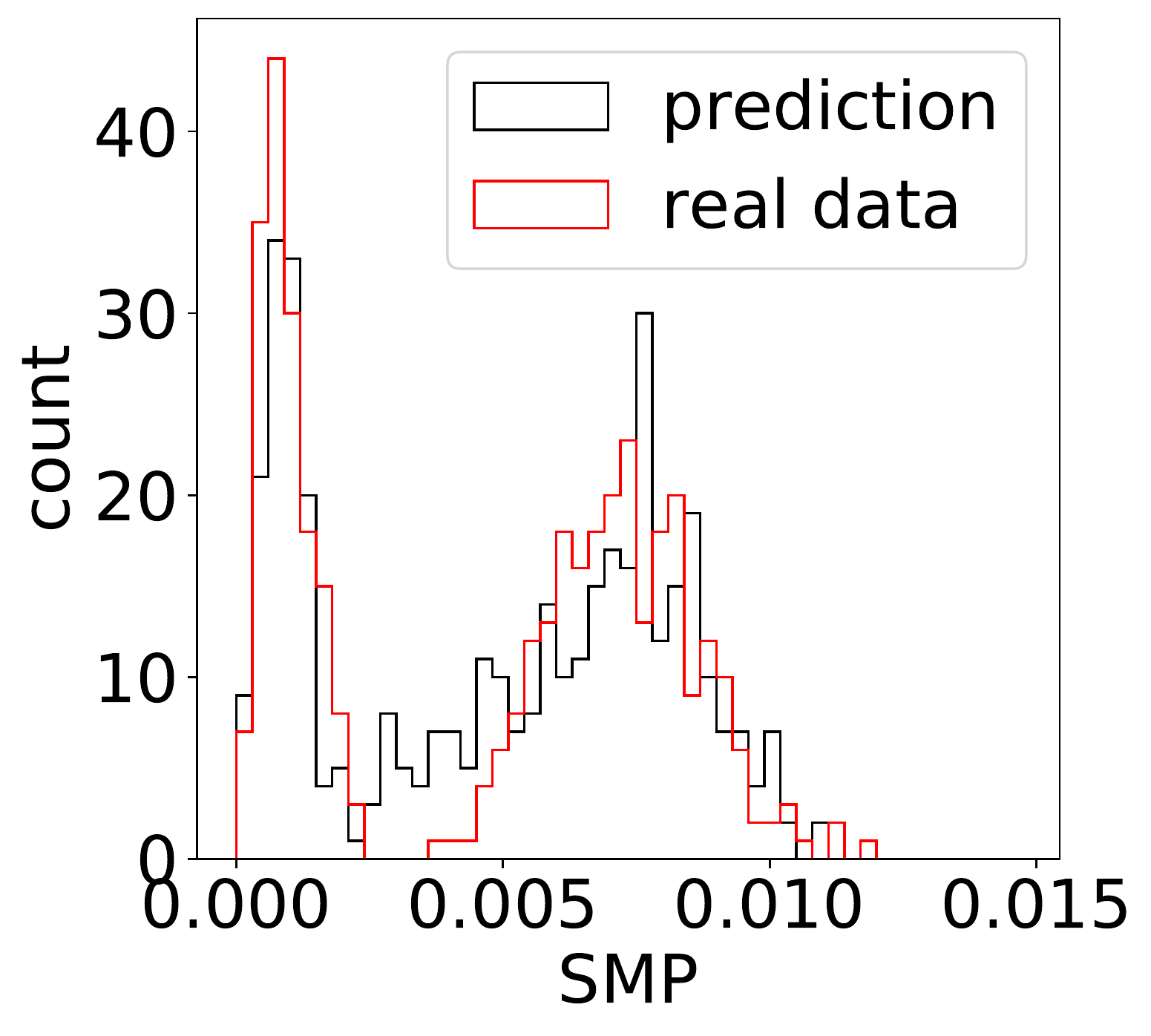}}
	
	\caption{The scatter plots for real and predicted SMPs with (a) a constant $w_0$ as the expectation of weight parameters and (b) the actual weights $W_{ij}$ in the trained autoencoders. The linear relationship between modeled and simulated results indicates the nonsmoothness events propagate through convolutional layers well. The scatter plots of SMP before and after (c) ReLU and (d) max pooling operations, the dashed line has a slope of one. (e) The distributions of predicted SMP and SMP in real data, the predicted distribution is consistent with the real data.}
	\vspace{-0mm}
	\label{fig:model_conv}
\end{figure*}

For a pixel location $(m,n)$ in the output, we used the model (\ref{conv_model_expectation}) to predict the SMP using the SMPs in input nodes, i.e., $\Tilde{\mu}^Y_{mn} = w_0 C_{\text{in}}\sum_u\sum_v \hat{\mu}^X_{mn} = w_0 \sum_u\sum_v\sum_c x_{cuv}$. We compared predicted SMPs $\Tilde{\mu}^Y$ and SMPs in real data $\hat{\mu}^Y$ for the nodes in the output, as shown in Fig.~\ref{fig:model_conv}(a).
We fitted a linear model using the SMPs in real data as the response and predicted SMPs as the predictor, the R-squared value was $0.842$. The high R-squared value indicates the model fits the nonsmoothness events well for the convolutional layer.
To further confirm the effectiveness of the model, we plugged in the actual weight parameters in the model instead of using $w_0$, i.e., $\Tilde{\Tilde{\mu}}^Y_{mn} = \sum_u\sum_v\sum_c |W_{cij}| x_{cuv}$. The comparison of predicted and real-data results are shown in Fig.~\ref{fig:model_conv}(b).  We fitted a linear model using the  simulation results as the response and modeled results as the predictor, and the R-squared value reached $0.944$.  
We note that using the expected weight $w_0$ in the model (\ref{conv_model_expectation}) can lower the modeling accuracy. However, in a real-world scenario, we need to use the expected weight because the individual weights of neural networks are usually unknown.
The confirmed linear relationship indicates that the nonsmoothness will propagate through the convolutional layers in a linear and almost deterministic way.

\subsection{Modeling of ReLU} \label{sec:model_relu}
ReLU activation will create or eliminate nonsmoothness events. 
The statistical effect of ReLU operation on the SMP is summarized in Fig.~\ref{fig:model_conv}(c) in the form of a highly structured joint distribution. The SMP can be reduced by a ReLU operation due to the deactivation of the node. The ReLU operation may also introduce additional nonsmoothness and therefore increasing the SMP. To predict SMP $y$ at the output of the ReLU activation given input SMP $x$, we observe from Fig.~\ref{fig:model_conv}(c) the concentrated density along the $x$-axis, the $y$-axis, and the $y=x$ line and therefore propose the following Bernoulli--Gaussian channel/conditional model: 
\begin{equation}
    f(y|x) = (1 - \theta) \cdot \delta(y) + \theta \cdot \mathcal{N}(y;x,\sigma),
\end{equation}
\noindent where $\theta$ is the parameter for Bernoulli distribution, $\delta$ is the Dirac delta function, and $\mathcal{N}(x;\mu,\sigma) = e^{-(x-\mu)^{2} /\left(2 \sigma^{2}\right)} \big/(\sigma \sqrt{2 \pi})$. The Bernoulli parameter $\theta$ and standard deviation $\sigma$ of Gaussian can be estimated from paired input--output data.

\subsection{Modeling of Max Pooling Layer} \label{sec:model_maxpool}
We also compared the SMPs in the input and output of the max pooling layers. For each pixel location in the output of max pooling layer, we calculated its SMP and compared it with the max value of SMPs in the corresponding locations in the input, as shown in Fig.~\ref{fig:model_conv}(d). The max pooling can introduce nonsmoothness and therefore increasing SMP, whereas it can also reduce SMP in the output by choosing a channel with no nonsmoothness events in the input. 
To predict the SMP at the output of the max pooling layer, we built a (simplified) conditional model for the max pooling layer based on the joint distribution in Fig.~\ref{fig:model_conv}(d). 
The output SMP $y$ follows a Gaussian distribution when input SMP $x$ is less than $a = 0.0025$, otherwise the output SMP $y$ follows a mixture of two Gaussian distributions:
\begin{equation}
\begin{split}
    & f(y|x) = \mathbbm{1}(x<a)\cdot \mathcal{N}(y;\mu_0,\sigma_0) + \mathbbm{1}(x\geq a)  \\ 
    & \cdot \left[ \pi_0 \mathcal{N}(y;x,\sigma_1) + (1-\pi_0)\mathcal{N}(y;\mu_2, \sigma_2) \right],
    \label{eq:maxpool_model}
\end{split}
\end{equation}
\noindent where $\mathbbm{1}$ is the indicator function.
The parameters of the Gaussian distributions $\mu_0, \sigma_0, \sigma_1, \mu_2$, $\sigma_2$ and the coefficient $\pi_0$ can be estimated from paired input--output data.

\subsection{Prediction of Nonsmoothness Events}
Finally, we predict the propagation of nonsmoothness events in a sequence of operations, i.e., the convolutional layer, ReLU activation, and the max pooling layer. Given the SMP to the input to the convolutional layer, we predict the SMP at the output of the max pooling layer with the linear model for the convolutional layer, the Bernoulli--Gaussian model for the ReLU, and the Gaussian mixture model for the max pooling layer. 
The distribution of predicted SMPs and SMPs in the simulated real data is compared in Fig.~\ref{fig:model_conv}(e). It shows that the distribution of prediction matches well with the real data.
Note that even though in (\ref{conv_model_expectation})--(\ref{eq:maxpool_model}) we have used parametric channel models, Monte-Carlo based methods can be used to address more generic channel models for nonsmoothness events.
In future work, we plan to conduct the joint analysis of all convolutional layers, transpose convolutional layers, max pooling layers, and the ReLU activation for characterizing the nonsmoothness of the neural network as a processing unit.
Such a statistical characterization of the processing unit can potentially enable forensic analysis of regression-based applications of neural networks.

\section{Conclusion}\label{sec:conclusion}
In this work, we have shown through synthetic data that modern neural networks using ReLU and max pooling can cause nonsmoothness. 
We have also modeled the nonsmoothness events in the input and output of different building blocks of neural networks.
To the best of our knowledge, this is the first work to investigate this understudied characteristic of the neural network, and we believe in its potential use as a generic forensic tool for regression-based neural network applications. 

\normalsize
\appendix
\subsection{Sequence of Images Forming a Smooth Trajectory in Euclidean Space}\label{sec:mnist}
\begin{figure}[!t]
    \centering
		\vspace{-3mm}
    \subfloat[]{\includegraphics[width=0.47\linewidth]{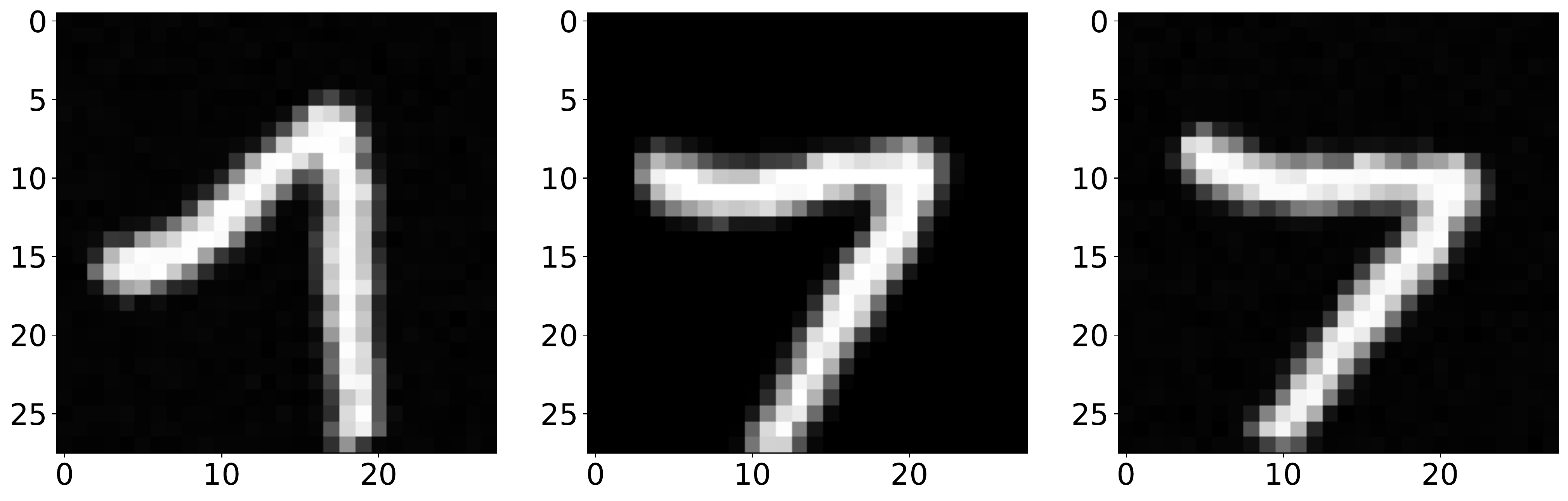}}
    \hspace{4mm}
    \subfloat[]{\includegraphics[width=0.47\linewidth]{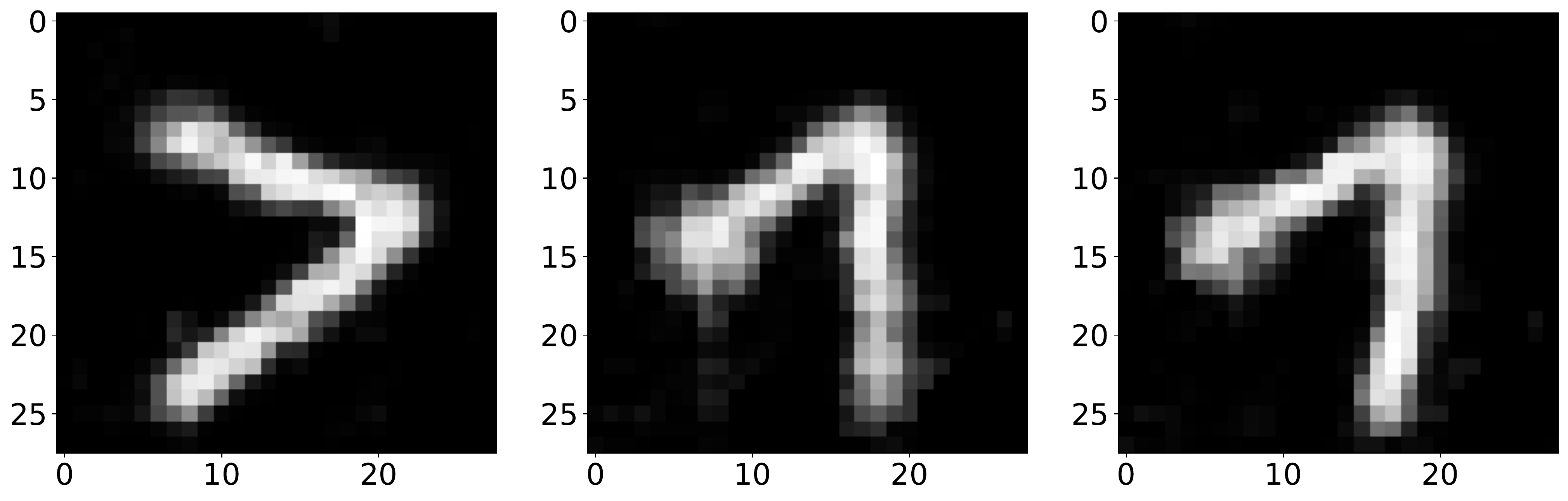}}
		\vspace{-0mm}
    \caption{(a) Three typical input images to the trained autoencoder, and (b) the corresponding reconstructed images. The reconstructed images are slightly blurred due to the encoding-decoding process.  }
		\vspace{-0mm}
    \label{minist_frames}
\end{figure}

\noindent \textbf{Dataset Generation and Autoencoder Training}\hspace{3mm}
In this simulation, we let the image vary smoothly in a Euclidean space instead of on a manifold.
We used digit ``7" images, a total of $6265$ images, from the MNIST dataset \cite{lecun2010mnist} to synthetically generate a smooth video dataset. To construct a training dataset, we rotated each image by $60$ randomly generated angles,  $\theta_i \overset{\text{iid}}{\sim}  \mathcal{U}(-30^{\circ}, 30^{\circ})$, where $\mathcal{U}$ denotes the uniform distribution. The total number of images in the training dataset was $6265 \times 60 = 375900$. Three typical images in the dataset are shown in Fig.~\ref{minist_frames}(a). 
To construct a validation dataset, we synthetically generated another $60$ angles for each of the $6265$ template images. 
        
We trained two autoencoders that can reconstruct the images in the synthetic dataset. 
One autoencoder follows Setup~1: ReLU activation and max pooling, and the other autoencoder follows Setup~2: softplus activation and average pooling. 
Since the image size is the same as the ellipsoid images, we used the same training set as in Section~\ref{sec:simulation} of the main paper.
The reconstructed images of Fig.~\ref{minist_frames}(a) using the trained autoencoder are shown in Fig.~\ref{minist_frames}(b).

We constructed a smooth video by adding a linearly increasing Gaussian noise image to a template image. Given a template image $\I_0$ in the training dataset and a noise image $\I_e$ of the same size, where each pixel value was independently drawn from the Gaussian distribution $\mathcal{N}(0, 1)$, we constructed a smooth video $\I(t)$ as follows:
\begin{equation}
    \I(t) = \I_0 + \alpha(t) \cdot \I_e,
    \label{eq:add_e}
\end{equation}
\noindent where $\alpha(t) = 10^{-2} (0.02 t - 1), t = 1,2,\dots, 100$, which is linearly increasing in time. $\I(t)$ is smooth because it is differentiable. This video can be regarded as image $\I_0$ propagating smoothly at a constant speed in an Euclidean space.
To obtain statistically confident results, we additionally obtained more realizations of autoencoders.
Using the same training and validation data, we repeated the training process to obtain ten autoencoders with \textsc{ReLU+MaxPool}, and another ten with \textsc{Softplus+AvePool}.
We used ten template images to construct ten smooth videos based on (\ref{eq:add_e}).
The ten smooth videos were reconstructed by all realizations of autoencoders, leading to a total of $10\times10 = 100$ reconstructed videos for \textsc{ReLU+MaxPool}, and another $100$ for \textsc{Softplus+AvePool}.

\begin{figure}[!t]
    \centering
		\vspace{-4mm}
    \subfloat[]{\includegraphics[width=0.49\linewidth]{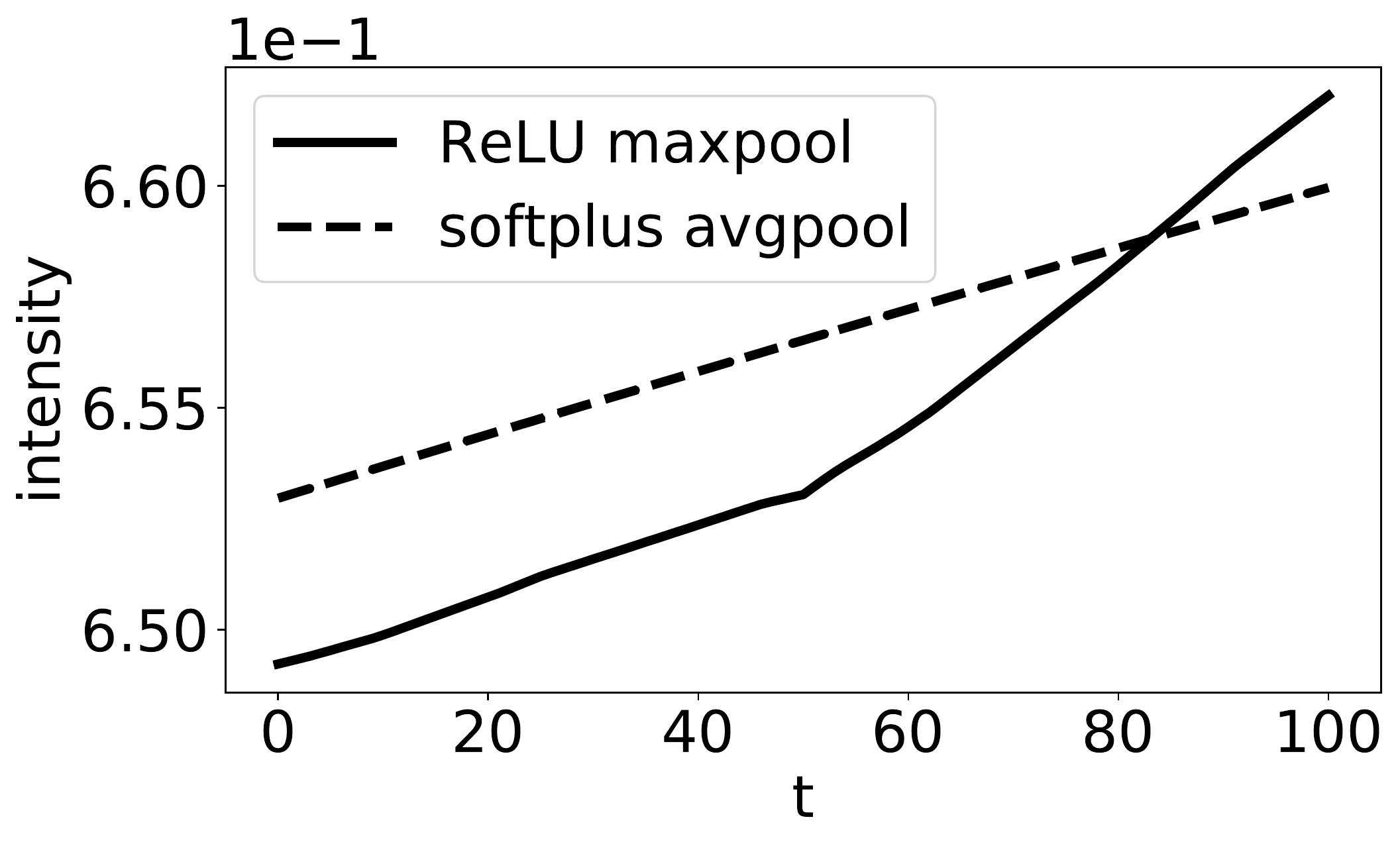}}
    \vspace{-0mm}
    \subfloat[]{\includegraphics[width=0.468\linewidth]{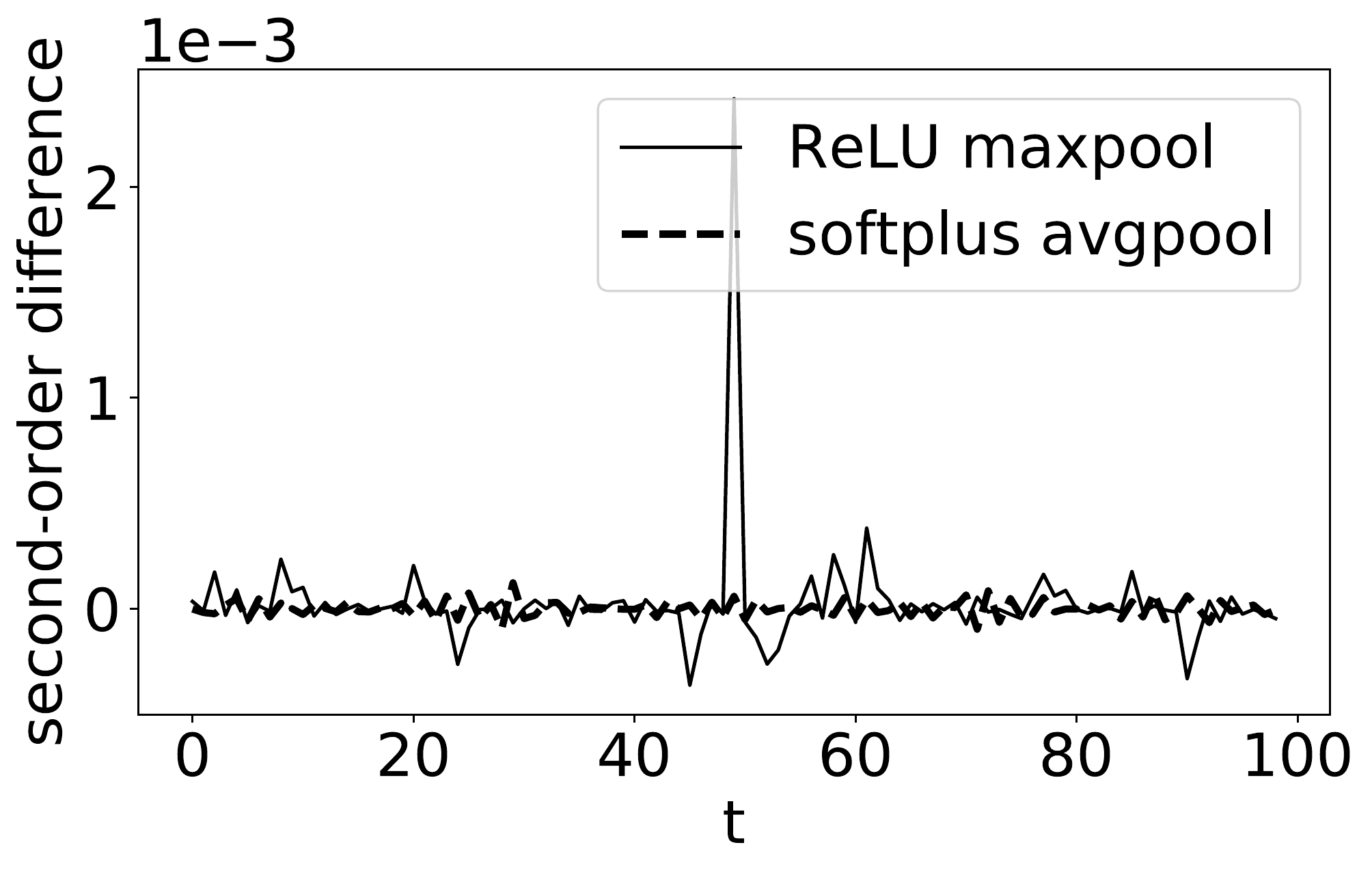}}
		\vspace{-0mm}
    \caption{For a representative pixel location, (a) the intensity curves, and (b) the second-order difference curves of reconstructed videos. The autoencoder with \textsc{ReLU+MaxPool} leads to larger second-order differences than the autoencoder with \textsc{Softplus+AvePool}, so it should have caused more nonsmoothness. 
    }
    \label{mnist_curves}
\end{figure}

\vspace{1mm}\noindent \textbf{Processing Smooth Input Videos by Neural Networks}\hspace{3mm}
For a smooth video $\I(t)$, we input frame by frame to a trained autoencoder to obtain the reconstructed video. 
For a pixel location, we compared the intensity time series of the reconstructed videos $\hat{\I}_{\text{ReLU+Max}}(t)$ and $\hat{\I}_{\text{Softplus+Ave}}(t)$ obtained using autoencoders with \textsc{ReLU+MaxPool} and \textsc{Softplus+AvePool}, respectively, as shown in Fig.~\ref{mnist_curves}(a). The second-order difference of the intensity time series were also calculated, as shown in Fig.~\ref{mnist_curves}(b). The autoencoder with \textsc{ReLU+MaxPool} caused larger peaks in the second-order difference curve than \textsc{Softplus+AvePool} given the same input video, confirming the nonsmoothness caused by \textsc{ReLU+MaxPool}.

\begin{figure}[!t]
    \centering
		\vspace{-4mm}
	\subfloat[]{\includegraphics[width=0.49\linewidth]{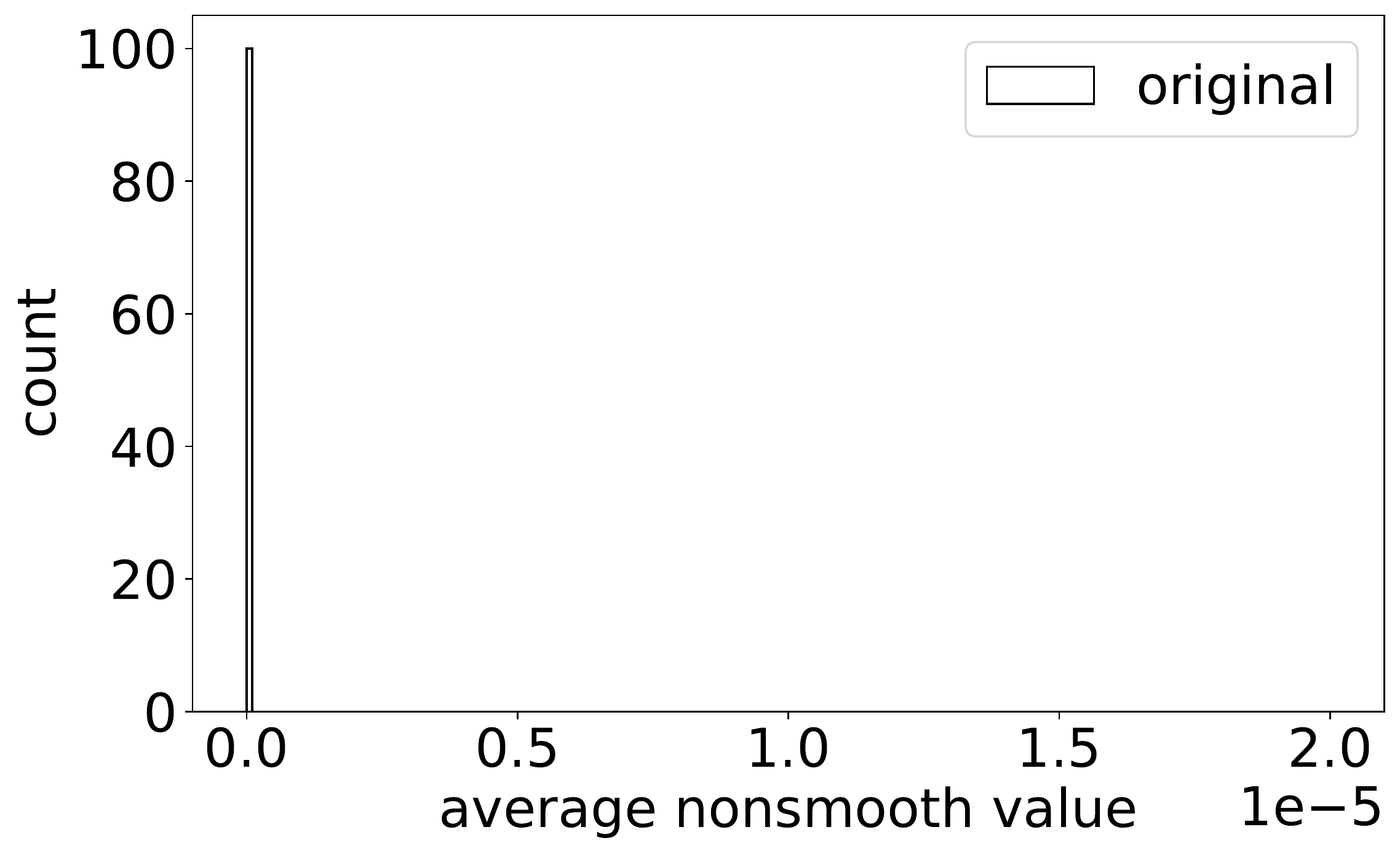}}
    \subfloat[]{\includegraphics[width=0.49\linewidth]{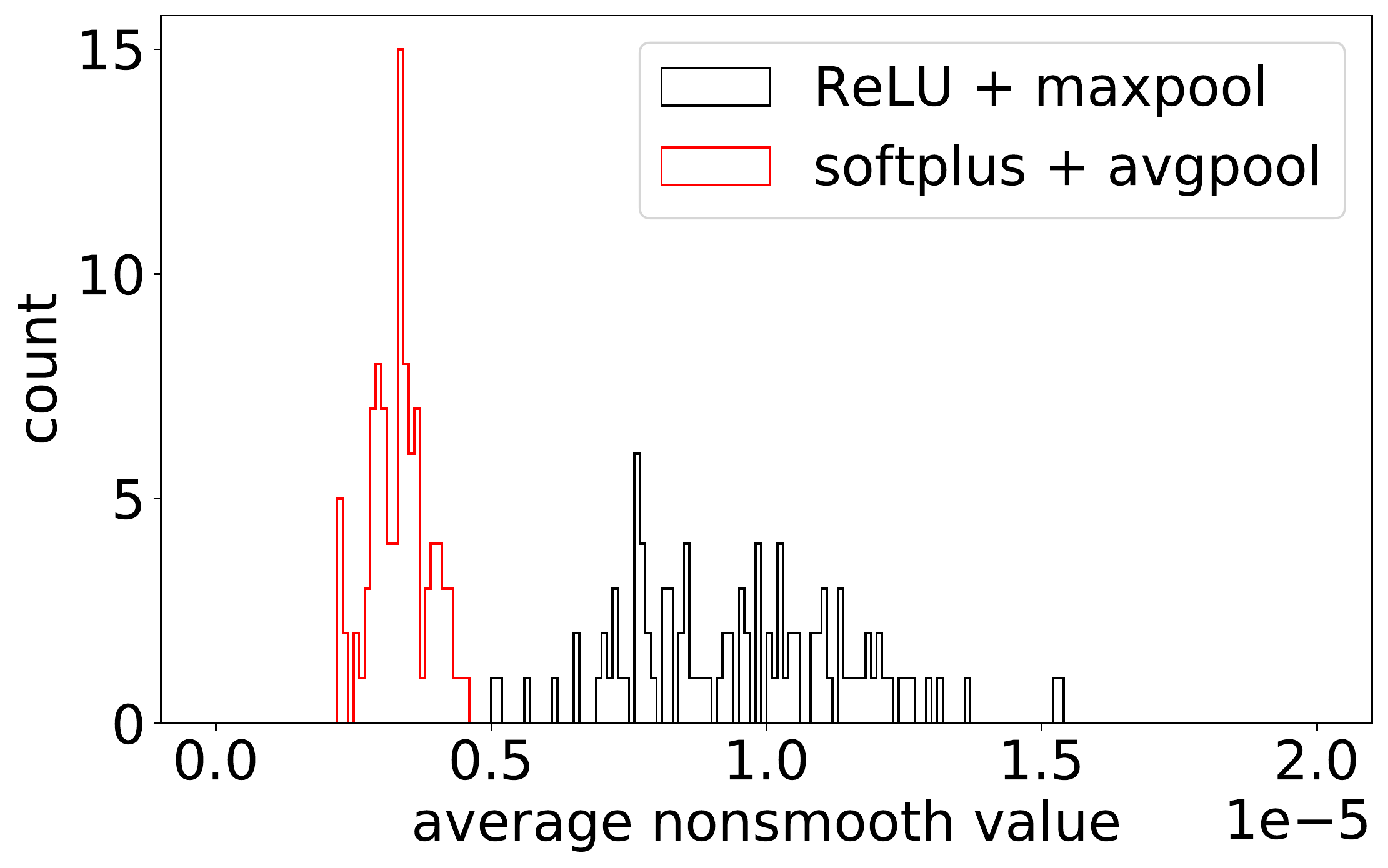}}
    \caption{The histograms of the AveNonSmooth of the (a) original input videos and (b) reconstructed videos. The AveNonSmooth is zero for the original input videos. When using autoencoder with \textsc{ReLU+MaxPool} to reconstruct videos, the AveNonSmooth are larger, indicating the nonsmoothness is introduced by \textsc{ReLU+MaxPool}. }
    \label{mnist_stat_res}
\end{figure}

We  examined  the  distribution of AveNonSmooth of original and reconstructed videos using autoencoders with different activation functions and pooling methods, as shown in Fig.~\ref{mnist_stat_res}. Both autoencoders raised AveNonSmooth in the reconstructed videos, but \textsc{ReLU+MaxPool} raised much more than \textsc{Softplus+AvePool}.
The slight increase of AveNonSmooth in the \textsc{Softplus+AvePool} setup can be attributed to the false positives due to the noisy reconstructed visual content, but it does not change the conclusion that \textsc{ReLU+MaxPool} leads to the nonsmoothness.

\subsection{Modeling Transpose Convolutional Layers}\label{sec:model_transconv}
We also model the nonsmoothness events in the transpose convolutional layers in the autoencoder.
Similar to a convolutional layer, an SMP in the output of a transpose convolutional layer can be expressed in a weighted summation of SMPs in the input.
We predicted the SMPs in the nodes of the three transpose convolutional layers of the autoencoders trained in Section~\ref{sec:simulation} of the main paper. 
We compared the predicted and real data, and the R-squared values for the fitted linear models for the three transpose convolutional layers are $ 0.79, 0.65,$ and $0.87$, respectively.
The large R-squared values indicate a linear relationship of SMPs between input and output of transpose convolutional layers.

\bibliographystyle{IEEEtran}
\bibliography{refs}


\end{document}